\documentclass[10pt,twocolumn,letterpaper]{article}

\usepackage[pagenumbers]{cvpr} % To force page numbers, e.g. for an arXiv version

\usepackage[dvipsnames]{xcolor}

\usepackage{algorithm}
\usepackage{algorithmic}
\usepackage{tabularray}
\usepackage{subcaption}
\usepackage[utf8]{inputenc}
\usepackage{booktabs}
\usepackage{tabularx}
\usepackage{threeparttable}
\usepackage{amsmath}
\usepackage{amssymb}
\usepackage{xcolor}
\usepackage{pifont}
\usepackage{multirow}

\definecolor{cvprblue}{rgb}{0.21,0.49,0.74}
\usepackage[pagebackref,breaklinks,colorlinks,citecolor=cvprblue]{hyperref}

\def\paperID{9} % *** Enter the Paper ID here
\def\confName{CVPR}
\def\confYear{2024}

\title{LD-Pruner: Efficient Pruning of Latent Diffusion Models \linebreak using Task-Agnostic Insights}

\author{Thibault Castells \and Hyoung-Kyu Song \and Bo-Kyeong Kim \and Shinkook Choi\\
Nota AI\\
Seoul, Korea\\
{\tt\small \{thibault, hyoungkyu.song, bokyeong.kim, shinkook.choi\}@nota.ai}
}

\begin{document}
\maketitle

\begin{abstract}
    Latent Diffusion Models (LDMs) have emerged as powerful generative models, known for delivering remarkable results under constrained computational resources. However, deploying LDMs on resource-limited devices remains a complex issue, presenting challenges such as memory consumption and inference speed. To address this issue, we introduce LD-Pruner, a novel performance-preserving structured pruning method for compressing LDMs. Traditional pruning methods for deep neural networks are not tailored to the unique characteristics of LDMs, such as the high computational cost of training and the absence of a fast, straightforward and task-agnostic method for evaluating model performance. Our method tackles these challenges by leveraging the latent space during the pruning process, enabling us to effectively quantify the impact of pruning on model performance, independently of the task at hand. This targeted pruning of components with minimal impact on the output allows for faster convergence during training, as the model has less information to re-learn, thereby addressing the high computational cost of training. Consequently, our approach achieves a compressed model that offers improved inference speed and reduced parameter count, while maintaining minimal performance degradation. We demonstrate the effectiveness of our approach on three different tasks:  text-to-image (T2I) generation, Unconditional Image Generation (UIG) and Unconditional Audio Generation (UAG). Notably, we reduce the inference time of Stable Diffusion (SD) by 34.9\% while simultaneously improving its FID by 5.2\% on MS-COCO T2I benchmark. This work paves the way for more efficient pruning methods for LDMs, enhancing their applicability.
\end{abstract}

\begin{figure}[thb!]
  \centering
  \includegraphics[width=1.0\linewidth]{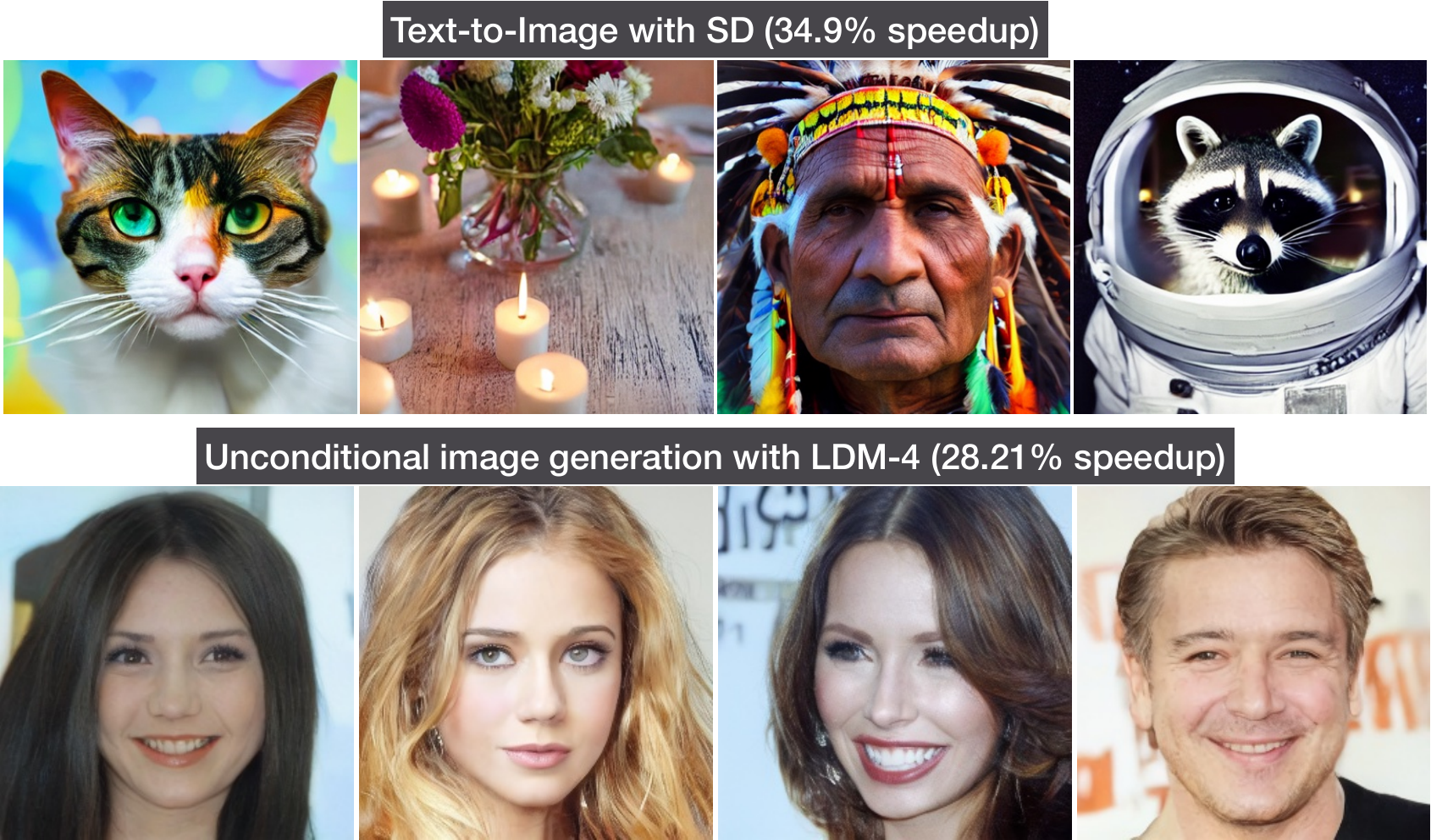}
  \caption{Samples generated using our compressed models. The proposed compression technique applies structured pruning to LDMs using task-agnostic information. Prompts (left to right): ``A multi-colored cat with yellow eyes staring upward'', ``Candles and flowers neatly placed on a table'', ``Portrait of a chief indian, 4k, high definition'', ``A photo of a raccoon wearing an astronaut helmet, looking out of the window at night.''}
  \label{fig:visual_results}
\end{figure}
\section{Introduction}
\label{sec:intro}

Generative models~\cite{gans, vae}, which can learn a data distribution and generate a sample from it, have revolutionized numerous domains, such as computer vision and natural language processing. Among them, Diffusion Models (DMs)~\cite{DDPM} have recently gained significant attention for their ability to generate high-quality images. LDMs, a subset of DMs that performs the diffusion process in a latent space, have witnessed rapid growth in popularity due to their fast generation capabilities and reduced computational cost~\cite{rombach2021highresolution, liu2023audioldm, zeng2022lion}. However, their deployment on resource-limited devices remains a challenge, mainly because of large compute requirements from the Unet in LDMs~\cite{BKsdm}.

A variety of strategies have been developed to compress LDMs and enhance their deployment feasibility, including quantization~\cite{li2023qdiff}, low-rank filter decomposition~\cite{svdiff}, and token merging~\cite{ToMeSD}. The primary goal of these techniques is to reduce the model's compute cost while striving to maintain its original performance, a crucial aspect of deploying models in resource-constrained environments. The work presented in this paper takes a distinct, yet complementary, approach to these studies.

Pruning, another compression technique which is traditionally utilized for the compression of convolutional networks by eliminating non-critical connections~\cite{l1norm, liunetworkslimming, AutoBot}, has been recently applied to DMs in the form of Diff-Pruning~\cite{fang2023depgraph}. This method identifies non-contributory diffusion steps and important weights using informative gradients, and applies filter pruning, significantly reducing computational overhead. However, Diff-Pruning does not extend its application beyond UIG. Moreover, its adaptability is further curtailed by the necessity to tune a threshold hyper-parameter for determining the optimal number of steps, as the ideal threshold is found to differ across datasets.

Training an LDM from scratch is both computationally demanding and financially costly. For instance, the reported training time for SD is a staggering 150,000 A100 hours~\cite{SD_training}, translating to an estimated cost on the order of magnitude of \$100,000. To ensure the retention of model performance throughout the pruning process, thus making the training faster, one might consider assessing the impact of pruning on model performance without fine-tuning~\cite{aimet}. While this method may prove effective for straightforward evaluative tasks like image classification, it becomes burdensome for generative models. Firstly, each different task—be it image generation, audio generation, and so on—requires its own unique evaluation tool. This introduces a lack of generalizability that compounds the complexity. Secondly, the performance evaluation of generative models is both complex and resource-intensive. Take, for example, the Fréchet Inception Distance (FID)~\cite{FID}, a commonly used evaluation metric for image generation. Its application requires the generation of thousands of images, a process that could take over an hour, making it impractical to use this metric for assessing the impact of each potential pruning operation. Moreover, the reliability of such metrics can be contentious~\cite{imagen}, further complicating the process.

We introduce a novel task-agnostic metric to measure the importance of individual operators, which are fundamental building blocks of the LDM architecture, such as convolutional layers and attention layers. We leverage this metric for structured pruning of LDMs. Our approach distinguishes itself by leveraging the latent space during the pruning process, specifically by assessing the impact of modifications within the model's latent representations. Operating in the latent space, where data is compact, provides dual benefits. Firstly, it ensures our method's independence from output types, facilitating a seamless adaptation to any task without necessitating adjustments. The use of cross-attention in the Unet of conditional LDMs to blend embeddings of different tasks in the latent space serves as a prime illustration of this task-agnostic property. Secondly, it yields computational efficiency, thereby addressing the performance evaluation challenge encountered in previous works.

Our method effectively identifies and removes components that contribute minimally to the output, leading to compressed models with faster inference speed and fewer parameters, without a major drop in performance. Through this work, we hope to extend the current body of work on LDM compression and enhance the deployment of LDMs in resource-constrained environments, expanding their applicability across various scenarios.

The main contributions of this paper are:

\begin{itemize}
    \item We propose a novel, comprehensive metric designed specifically to compare the latent representations of LDMs. This metric is underpinned by thorough experimental evaluations and logical reasoning, ensuring that each element of its design contributes effectively to the accurate and sensitive comparison of LDM latents.
    \item Leveraging this new metric, we formulate a novel, task-agnostic algorithm for compressing LDMs through architectural pruning. The primary focus of our proposed method is to maintain output quality during the pruning process, thereby accelerating the finetuning phase as the weights are preserved.
    \item We demonstrate the versatility of our approach through its application in three distinct tasks: T2I generation, UIG and UAG. The successful execution of these experiments underscores our method's potential for wide applicability across diverse tasks.
\end{itemize}

The remainder of this paper is organized as follows: Sec.~\ref{sec:method} presents our proposed method in detail, highlighting its novelty and how it overcomes the aforementioned limitations. Sec.~\ref{sec:experiments} outlines the experimental setup and evaluation metrics. Finally, Sec.~\ref{sec:discussion} compares our approach to existing methods and provides further analysis of our design choices and discusses potential areas for future improvement.
% =================================================================
% ============================ METHOD =============================
% =================================================================

\begin{figure*}[t!]
  \centering
  \includegraphics[width=1.0\linewidth]{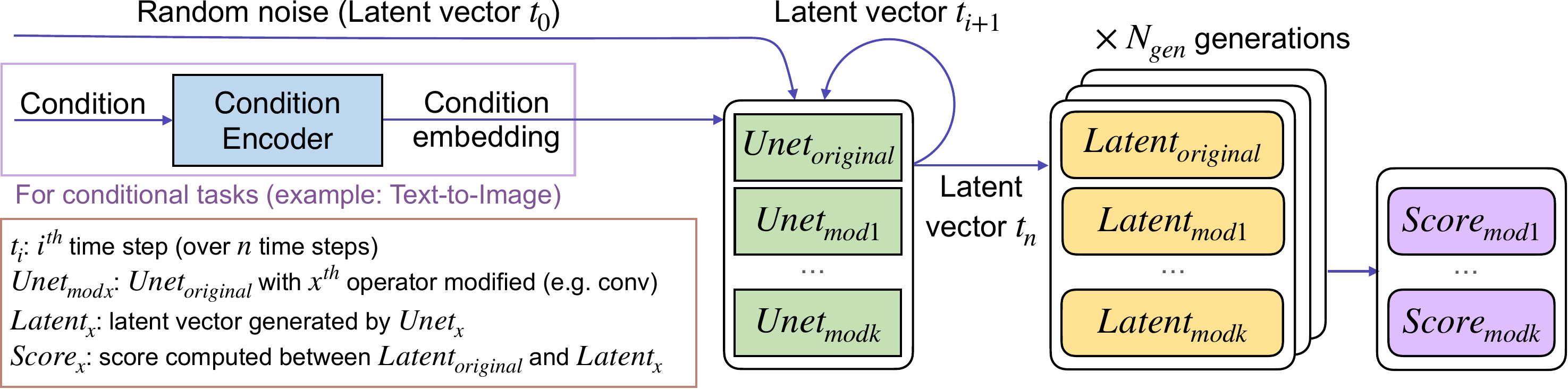}
  \caption{Overview of LD-Pruner. Given $k$ operators in the Unet, we generate $k+1$ sets of $N_{gen}$ latent vectors: one set for the original Unet, and one for each Unet where a single operator has been modified. The importance score of each operator is then calculated using a formula specifically designed to compare latent vectors. This formula, sensitive to shifts in both the central tendency and the variability of the latent vectors, generates a comprehensive measure of the importance of each operator.}
  \label{fig:pipeline}
\end{figure*}

\section{Proposed Method}
\label{sec:method}

This section describes our novel algorithm for compressing LDMs. The proposed method employs structured pruning aimed at minimizing performance loss, regardless of the generation task.

\subsection{Method Overview}

Our ultimate goal is to enhance an LDM efficiency by minimizing the presence of less impactful operators, thus streamlining the architecture without sacrificing effectiveness. In the preliminary stage of our approach, we focus on collecting the information needed to assess the individual significance of each operator within the Unet of the LDM. This involves systematically modifying each operator to simulate potential pruning effects, and carefully tracking these modifications by generating unique latent representations associated with each change. This information gathering enables us to determine the potential impact of removing or reducing certain components, guiding our subsequent pruning decisions. Using a tailored scoring formula, we quantify the divergence between the original and altered latent representations, offering insight into the impact of each modification. The scores derived from this process then inform our pruning phase, where we decide which operators to prune or replace. A visual depiction of this method, highlighting the key stages of the computation of our pruning score, is provided in Fig.~\ref{fig:pipeline}.

\subsection{Operator Modification and Latent Representation Collection}

Transitioning to the practical aspects of our methodology, we first focus on the meticulous modification of the operators within the Unet of the LDM. In the context of our work, an operator refers to a fundamental building block of the architecture, such as convolutional layers, attention layers, normalization layers, activation functions, or more complex components like transformer blocks. For each of these operators, we consider one of two possible modifications. First, we attempt to eliminate the operator in its entirety, provided this action is feasible. However, there are situations where operator removal is not viable—specifically, when the operator's input and output dimensions do not align.
In these instances, we resort to an alternative approach: replacing the operator with a less computationally demanding operation that retains the original dimensions. If there is a disparity in the number of channels, we use a 1$\times$1 convolution operation to match the dimensions without adding significant computational overhead. In cases where the spatial resolution varies, we use average pooling or upscaling. For implementation details, please refer to the Supplementary Materials.

Following the modification of an operator, we generate multiple latent representations using the modified model. Each set of latent representations corresponds to a specific operator modification and collectively forms a comprehensive record of the model's output under various operator modifications. After each set of latent representations is obtained, we restore the modified operator to its original state. This ensures that each operator is modified in isolation, preventing the cumulative effects of multiple modifications from influencing the assessment of individual operators.

For the original, unmodified model, we also generate latent representations (referred to as the original set) that serve as a baseline for comparison. This comprehensive collection of latent representations, both from the original and modified models, forms the foundation for the subsequent evaluation of operator significance in our methodology.

\subsection{Operator Significance Evaluation}

The evaluation of operator significance is a critical step in our pruning method. This process is based on the assumption that a significant change in the latent space is likely to lead to a substantial change in the model's output. In order to quantify this change and thus estimate the significance of each operator, we employ a specially designed scoring formula crafted to capture the difference between two sets of latent representations.

Let $\mathbf{L}_{orig}$ and $\mathbf{L}_{mod}$ denote the original set and a modified set, respectively, each of which contains $N_{gen}$ latent representations. We denote the $i$-th latent vector in $\mathbf{L}_{orig}$ and $\mathbf{L}_{mod}$ as $\mathbf{l}_{orig,i}$ and $\mathbf{l}_{mod,i}$, respectively. The scoring formula consists of two main components: the average distance, denoted as $avg_{dist}$, measures the distance between the average values of $\mathbf{L}_{orig}$ and $\mathbf{L}_{mod}$, and the standard deviation distance, denoted as $std_{dist}$, measures the distance between the standard deviations of $\mathbf{L}_{orig}$ and $\mathbf{L}_{mod}$. Formally, we compute $avg_{dist}$ and $std_{dist}$ as follows:

\begin{align}
    avg_{dist} &= |avg_{orig} - avg_{mod}|_2, \\
    std_{dist} &= |std_{orig} - std_{mod}|_2
\end{align}

\noindent where $|\cdot|_2$ denotes the Euclidean norm and where:

\begin{align}
    avg_{orig} &= \frac{1}{N_{gen}}\sum_{i=1}^{N_{gen}}\mathbf{l}_{orig,i}, \\
    avg_{mod} &= \frac{1}{N_{gen}}\sum_{i=1}^{N_{gen}}\mathbf{l}_{mod,i}
\end{align}

\noindent and:

\begin{align}
    std_{orig} &= \sqrt{\frac{1}{N_{gen}}\sum_{i=1}^{N_{gen}}(\mathbf{l}_{orig,i} - avg_{orig})^2}, \\
    std_{mod} &= \sqrt{\frac{1}{N_{gen}}\sum_{i=1}^{N_{gen}}(\mathbf{l}_{mod,i} - avg_{mod})^2}
\end{align}

The score $S$ for each operator is then computed by summing $avg_{dist}$ and $std_{dist}$:

\begin{equation}
    \text{score} = avg_{dist} + std_{dist}.
    \label{eq:score}
\end{equation}

\noindent  This scoring formula is designed to be sensitive to both shifts in the central tendency and changes in the variability of the latent representations, providing a comprehensive measure of the impact of operator modification. By using this formula, our method can effectively identify the operators that are most (higher score) and least (lower score) significant to the model's performance, guiding the pruning process toward the most efficient and least disruptive modifications. We further discuss our metric in Sec.~\ref{subsec:metric_justification}.

\begin{table*}[ht!]
    \scriptsize % Small font size
    \centering
    \begin{threeparttable}
        \begin{tabularx}{0.8\textwidth}{X@{\hskip3pt}|r@{\hskip3pt}|r@{\hskip3pt}|r@{\hskip3pt}|r@{\hskip3pt}|r@{\hskip3pt}|r}
            \toprule
            Model & FID $\downarrow$ & IS $\uparrow$ & CLIP $\uparrow$ & \# Params & Data Size & Speedup \\
            \midrule
            SD-v1.4~\cite{rombach2021highresolution}  & 13.05 & \textbf{36.76} & \textbf{0.2958} & 1.04B & $>$2000M & 0\% \\
            \hline
            \textbf{LD-Pruner (ours)} (42 modifications) & \textbf{12.37} & \textbf{35.77} & \textbf{0.2894} & \textbf{0.71B} & \textbf{0.22M} & \textbf{34.89\%} \\
            \hline
            Small Stable Diffusion~\cite{OFA_sdm} & 12.76 & 32.33 & 0.2851 & 0.76B & 229M & 35.28\% \\
            BK-SDM-Base~\cite{BKsdm} & 15.76 & 33.79 & 0.2878 & 0.76B & \textbf{0.22M} & 35.28\% \\
            BK-SDM-Small~\cite{BKsdm} & 16.98 & 31.68 & 0.2677 & 0.66B & \textbf{0.22M} & \textbf{36.98\%} \\
            \hline
            DALL·E~\cite{dalle} & 27.5 & 17.9 & - & 12B & 250M & \\
            DALL·E-2~\cite{dalle2} & \textbf{10.39} & - & - & 5.2B & 250M & \\
            CogView~\cite{cogview} & 27.1 & 18.2 & - & 4B & 30M & \\
            CogView2~\cite{cogview2} & 24.0 & 22.4 & - & 6B & 30M & \\
            Make-A-Scene~\cite{makeascene} & 11.84 & - & - & 4B & 35M & \\
            LAFITE~\cite{lafite} & 26.94 & 26.02 & - & \textbf{0.23B} & 3M & \\
            GALIP (CC12M)~\cite{galip} & 13.86 & 25.16 & 0.2817 & 0.32B & 12M & \\
            GLIDE~\cite{glide} & 12.24 & 30.29 & - & 5B & 250M & \\
            LDM-KL-8-G~\cite{rombach2021highresolution} & 12.63 & - & - & 1.45B & 400M & \\
            SnapFusion~\cite{snapfusion} & $\sim$13.6 & - & $\sim$0.295 & 0.99B & $>$100M & \\
            Würstchen-v2~\cite{wurstchen_v2} & 22.40 & 32.87 & 0.2676 & 3.1B & 1700M & \\
            \bottomrule
        \end{tabularx}
        \begin{tablenotes}
            \item For the IS and FID values of comparative models, we adopt the evaluations as reported in~\citet{BKsdm}
        \end{tablenotes}
        \caption{Comparison of different models for T2I Generation, on the MS-COCO $256\times256$ validation set. Speedup values are measured relatively to SD-v1.4.}
        \label{tab:t2i_comparison}
    \end{threeparttable}
\end{table*}

\subsection{Model Pruning}

After calculating the significance scores for each operator, the computed scores serve as a roadmap, guiding us in identifying the operators that could be pruned or substituted with the least potential impact on the model's performance. Our strategy particularly focuses on operators with the lowest scores for elimination, since these are regarded as the least contributory to the model's output.

Determining the number of operators to prune, denoted as $k$, requires a deliberate and systematic evaluation. This process is essentially a trade-off exercise between achieving model compression and preserving performance. An increase in the number of pruned operators leads to a more compact model, however, it may also risk a significant reduction in performance. Therefore, our goal is to identify an optimal value for $k$ that offers a substantial degree of model compression while maintaining satisfactory performance levels. Such trade-off is discussed in Sec.~\ref{subsec:tradeoff}.

In the case of conditional LDMs, such as SD, we conduct the evaluation for various conditions. For every operator, we aggregate the scores across all conditions. This approach ensures that the pruning does not overly specialize for a specific condition. In practice, we used 50 different prompts conditions for our SD experiment. The entire pruning process is concisely summarized in Alg.~\ref{alg:pruning_dm}. Notably, in this context, an unconditional task can essentially be interpreted as a conditional task with only a single implicit condition.

Upon completing the pruning process, we engage in fine-tuning the pruned model to recoup any performance reduction that occurred as a consequence of the pruning operation.

\begin{algorithm}[thb]
    \small
    \caption{Efficient Pruning for LDMs}
    \label{alg:pruning_dm}
    \textbf{Input}: LDM unet $Unet$, list of condition $CList$, generation per condition $N_{gen}$, number of operator to prune $k$ \\
    \begin{algorithmic}[1] %[1] enables line numbers
        \STATE $scores$ $\gets$ empty dictionary
        \FOR{$C$ in $CList$}
            \STATE $latent_{orig}$ $\gets$ $generate(Unet, C, N_{gen})$
            \FOR{operator $op$ in $Unet$}
                \STATE $Unet_{mod}$ $\gets$ $prune(Unet, op)$
                \STATE $latent_{mod}$ $\gets$ $generate(Unet_{mod}, C, N_{gen})$
                \STATE $s$ $\gets$ $get\_score(latent_{orig}, latent_{mod})$ (\text{Eq.~\ref{eq:score}})
            \STATE $scores[op.name]$ $\gets$ $scores[op.name] + s$
            \ENDFOR
        \ENDFOR
        \STATE $Unet$ $\gets$ $prune(Unet, scores, k)$
    \end{algorithmic}
\end{algorithm}

\subsection{Complexity Analysis}

We provide a time complexity analysis of our proposed algorithm, focusing on the main operations involved. Let's denote $n$ the number of computational operations in the Unet to generate a single latent representation, $m$ the total number of operators that are potential candidates for pruning, and $k$ the number of latent representations to generate for each operator modification. Given these definitions, our algorithm's time complexity can be expressed as $O(nmk)$. This complexity can be effectively managed in practice. For instance, we can decrease $m$ by filtering out operators that contribute negligibly to the overall model latency, thereby focusing our efforts on the more significant contributors. An additional advantage stems from the specific nature of our compression method, which operates in the latent space of the LDMs. In contrast to other pruning methods that require the generation of full outputs, our technique only needs the latent representations. This allows us to avoid the decoding step during generation, thereby reducing the overall computational burden and accelerating the pruning process.
\section{Experimental Setup}
\label{sec:experiments}
% =================================================================
% =================== EXPERIMENTAL RESULTS ========================
% =================================================================

To highlight the task agnostic property of the proposed importance score, we apply it to three different tasks: T2I Generation with SD-v1.4~\cite{rombach2021highresolution}, UIG with LDM-4~\cite{rombach2021highresolution} and UAG with AudioDiffusion~\cite{audiodiffusion}.

\subsection{Training}

For each task, we finetune our compressed Unets employing Knowledge Distillation (KD), applied both at the feature and output levels~\cite{BKsdm}. For the detailed hyper-parameters, please refer to the Supplementary Materials.

\noindent
{\bf T2I Generation.} We finetune our compressed model on a subset of $0.22$M image-text pairs from the LAION-Aesthetics V2 6.5+ dataset~\cite{laion_aesthetics}, which represents less than $0.1$\% of the training pairs used in the LAION-Aesthetics V2 5+~\cite{laion_aesthetics} for training SD-v1.4. All training is conducted on a single A100 GPU.

\noindent
{\bf Unconditional Image Generation.} For UIG, we leverage the complete CelebA-HQ $256\times256$ dataset~\cite{celeba_hq} due to its relatively small size (approximately $30$k images). All training is conducted on a single NVIDIA GeForce RTX 3090 GPU.

\noindent
{\bf Unconditional Audio Generation.} For UAG, we finetune using the same dataset that was employed to train AudioDiffusion. This dataset consists of 20k Mel spectrograms of size $256\times256$, generated from $5$-second audio files. All training is conducted on a single NVIDIA GeForce RTX 3090 GPU.

\subsection{Evaluation}

\noindent
{\bf Performance metric.} We report the FID as our main performance metric for image generation. In the case of T2I generation, the FID is measured by generating $30$k samples from the MS-COCO $256\times256$ validation set~\cite{COCO}. In Tab.~\ref{tab:t2i_comparison}, we additionally present the Inception Score (IS)~\cite{IS}, computed using the same dataset as for the FID computation. In the case of UIG, the FID is measured by generating $5$k samples, and we compute the FID with the training set as commonly done with CelebA-HQ~\cite{rombach2021highresolution}. For UAG, we employ the Fréchet Audio Distance (FAD)~\cite{FAD}, a specialized variant of the FID tailored for audio comparison. Again, the FAD is measured between $5$k generated samples (following the author's recommendation) and the training set.

\noindent
{\bf Computation Efficiency metric.} Unlike some previous work, such as Diff-Pruning, our study focuses on real-world inference speed improvements rather than relying solely on FLOPs and MACs, which can be unreliable predictors of actual speed~\cite{fernandez2023framework, li2021hwnasbench}. We conduct our inference speed evaluations for T2I generation and UIG on a single NVIDIA GeForce RTX 3090 GPU, performing $30$ and $200$ inference steps, respectively. For UAG, we utilize a CPU (Intel Xeon Silver 4210R) due to the model's small size, and carry out $50$ inference steps ($\sim 5$-$6$ seconds). To ensure the stability and reliability of our results, we perform a warmup process by initially generating $20$ samples. Subsequently, we compute the average speed over $100$ generated samples. This approach offers a more robust and realistic estimate of the model's operational efficiency in real-world applications.
\section{Results and Discussion}
\label{sec:discussion}
% =================================================================
% ======================== DISCUSSION =============================
% =================================================================

\begin{figure*}[thb!]
  \centering
  \includegraphics[width=1\linewidth]{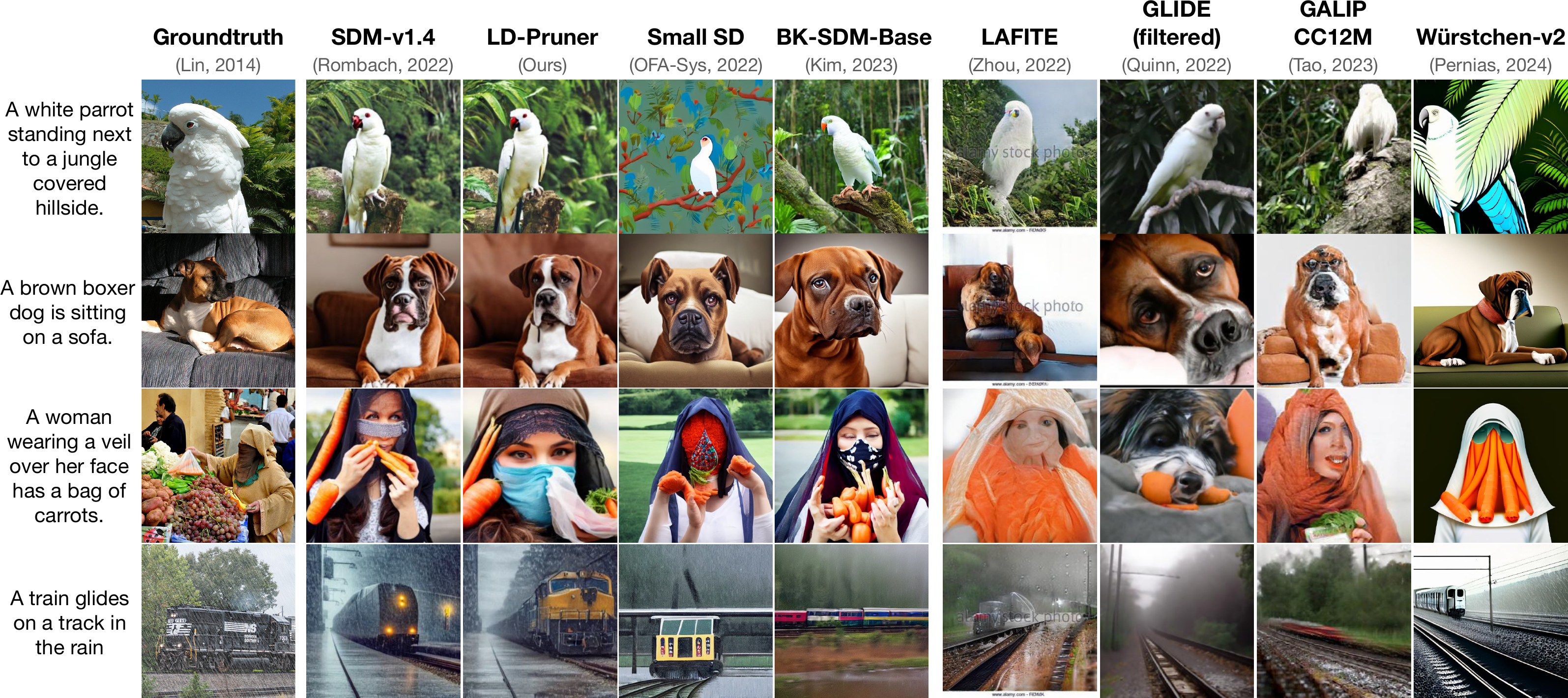}
  \caption{Qualitative comparison on zero-shot MS-COCO benchmark on T2I. The results of previous studies were obtained with their official released models.}
  \label{fig:T2I_qualitative_results}
\end{figure*}

\subsection{Main Results}
\label{subsec:results}

\begin{figure}[thb!]
  \centering
  \includegraphics[width=1.0\linewidth]{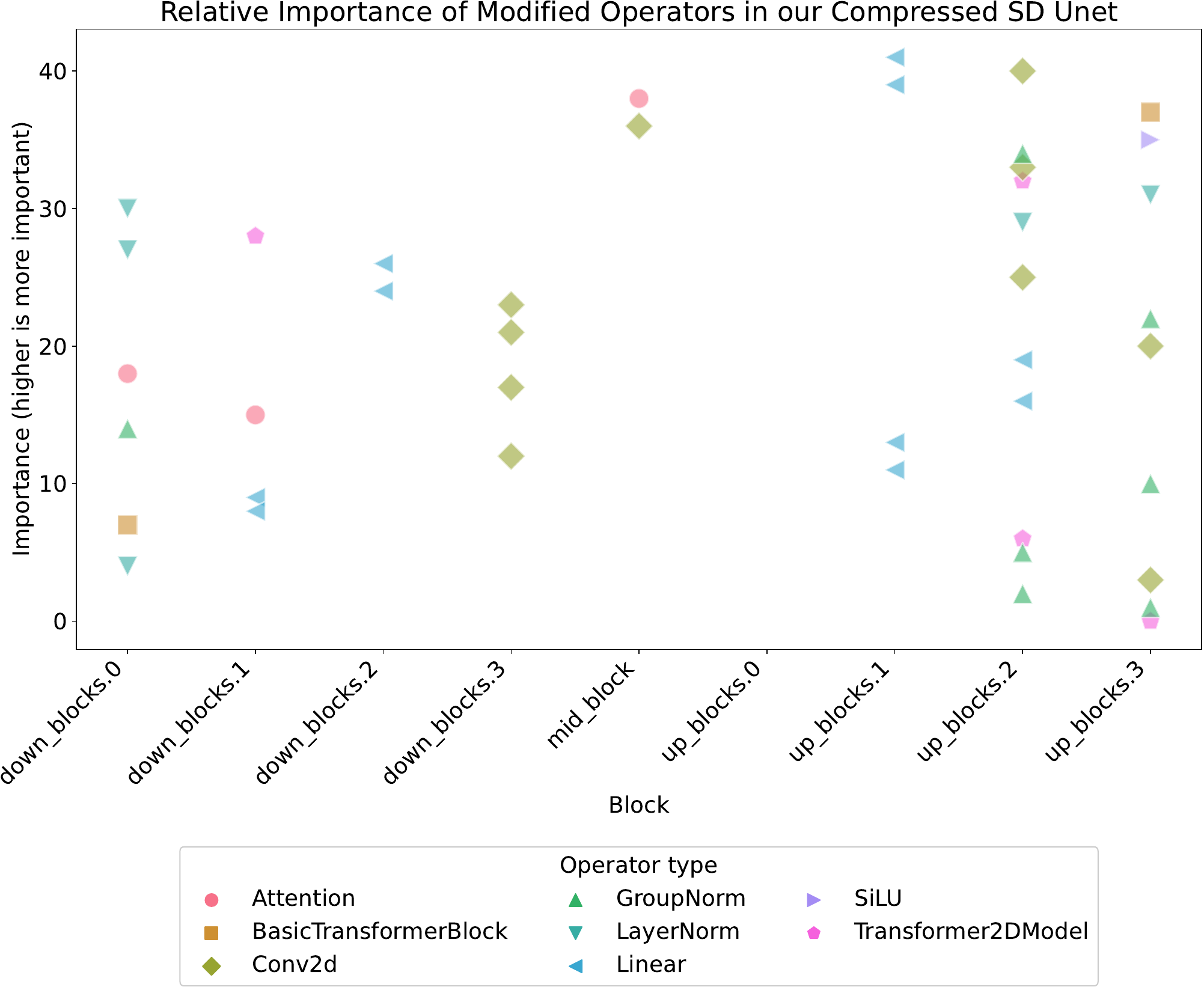}
  \caption{Type and relative importance of the modified operators in each block of our compressed SD.}
  \label{fig:block_vs_rank}
\end{figure}

\noindent
{\bf T2I Generation.} In Tab.~\ref{tab:t2i_comparison}, we benchmark our pruned SD models against other T2I models. Relative to the manual architectural pruning of SD-v1.4~\cite{rombach2021highresolution}, our model surpasses those finetuned with large data and does so with a remarkable reduction in the number of training data samples — $1040$ times fewer than used in~\citet{OFA_sdm}, a similar approach to that of~\citet{BKsdm}. A qualitative comparison can be found in Fig.~\ref{fig:T2I_qualitative_results}. Furthermore, our pruned models demonstrate competitive performance when juxtaposed with other architecture, including those based on autoregression~\cite{dalle, cogview, cogview2, makeascene}, GANs~\cite{lafite, galip}, and diffusion methods~\cite{glide, rombach2021highresolution, dalle2, wurstchen_v2}. Lastly, Fig.~\ref{fig:block_vs_rank} visualizes the modified operators and their importance rankings. Notably, a significant portion of these operators is located near the model's output, suggesting an over-parametrization in this region.

\begin{figure}[thb!]
  \centering
  \includegraphics[width=1\linewidth]{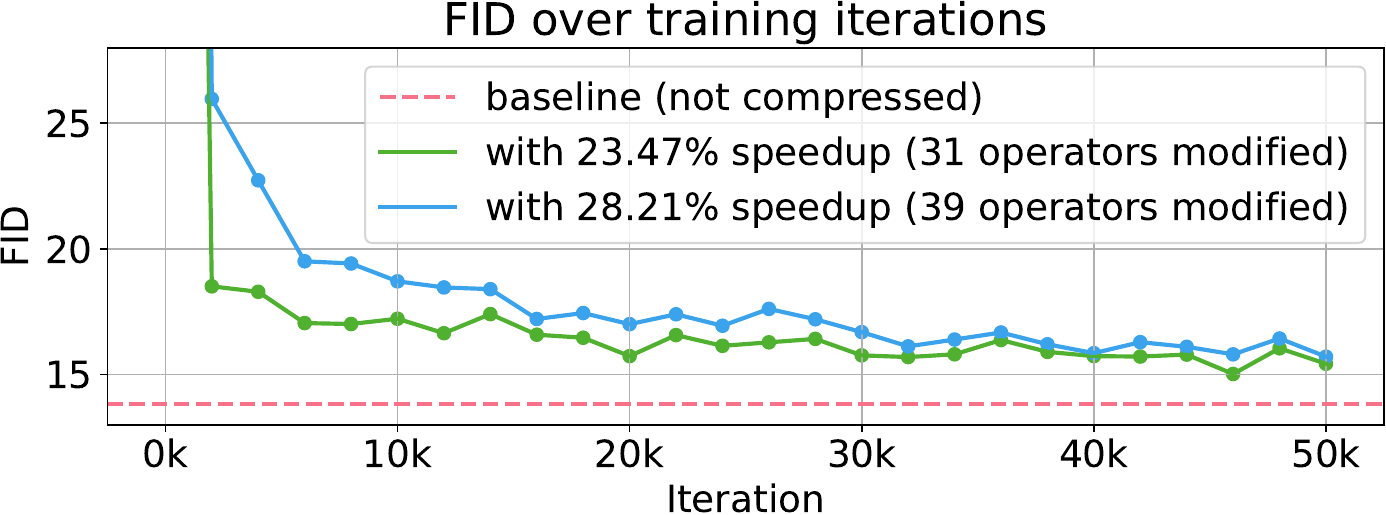}
  \caption{Evolution of the FID during the training process for the UIG task on the CelebA-HQ $256\times256$ dataset, for two different compression ratios.}
  \label{fig:fid_vs_iteration_uncond_ldm}
\end{figure}

\noindent
{\bf Unconditional Image Generation.} We show the evolution of the FID during training in Fig.~\ref{fig:fid_vs_iteration_uncond_ldm}. We modified $31$ operators resulting in a $23.47$\% speedup and $39$ operators for a $28.21$\% speedup. In both cases, the compressed model converges rapidly, reaching a minimum FID of $15.03$ after $46$k iterations and $15.71$ after $50$k iterations, respectively. For comparison, the original model reaches an FID of $13.84$ after $410$k iterations. Notably, with just $20$k iterations and $20$ modified operators (corresponding to an $18.23$\% speedup), we surpass the FID performance of the original model as shown in Fig.~\ref{fig:avg_std_quantitative}, achieving an FID of $13.72$. This underlines the efficiency of our approach in both speed and performance dimensions.

{\bf Unconditional Audio Generation.} Tab.~\ref{tab:audiodiffusion_comparison} showcases the comparison between our compressed model and the baseline AudioDiffusion model. With $90$ operators modified, we achieve a $19.2$\% speedup and a post-finetuning FAD of $2.0$ ($-0.5$). These results underscore LD-Pruner's ability to compress independently of the task at hand. It is worth mentioning that the audio output from the pruned model sounds the same as the audio of the original model. This can be quantified by evaluating the FAD between $5000$ samples of both models, resulting in a score of $0.05$.

\begin{table}[thb!]
    \small  
    \centering
    \begin{tabular}{l|c|c|c|c}
    \toprule
    Pruning & Finetuned & FAD $\downarrow$ & \# Params & Speedup \\
    \midrule
    None       & -                   & 2.3              & 163.1M    & 0\%  \\
    \midrule
    \multirow{3}{*}{LD-Pruner} & \textcolor{red}{\ding{55}} (reset Unet) & 13.4  & \multirow{3}{*}{85.6M}  & \multirow{3}{*}{19.2\%} \\
                               & \textcolor{red}{\ding{55}}   & 8.7   &                         &                         \\
                               & \textcolor{green}{\ding{51}} & \textbf{2.0}   &                         &                         \\
    \bottomrule
    \end{tabular}
    \caption{Compression performance on UAG task with AudioDiffusion. When finetuning, we proceed for 12k steps.}
    \label{tab:audiodiffusion_comparison}
\end{table}

\subsection{Scoring Metric Composition}
\label{subsec:metric_justification}

An essential aspect of our methodology is the scoring metric, as the pruning quality depends on it. Therefore, before settling on the chosen method, we carried out extensive experimentation with alternative ways of incorporating the statistical measures.

\begin{figure}[thb!]
  \centering
  \includegraphics[width=1\linewidth]{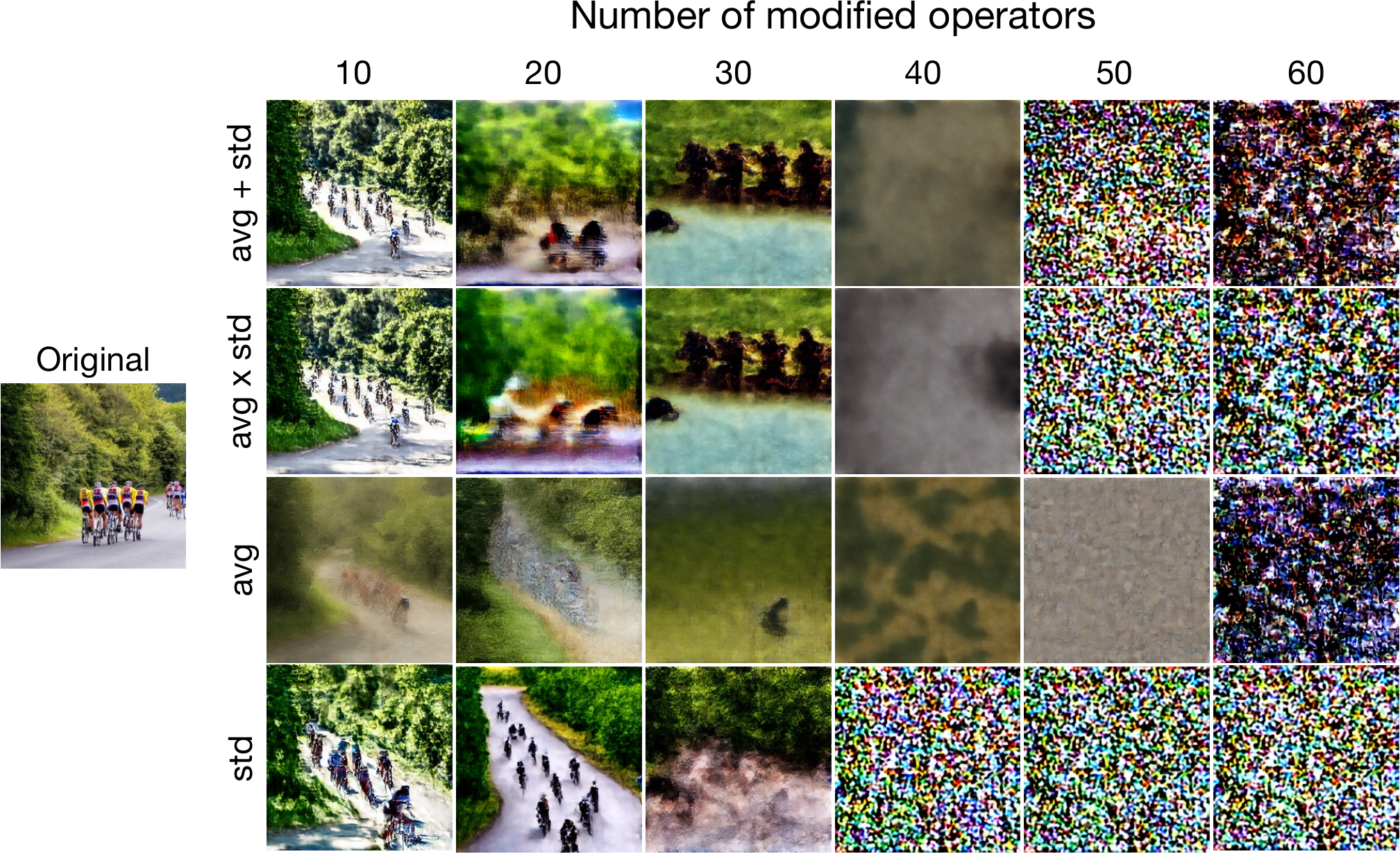}
  \caption{Qualitative comparison of the impact of various combination methods for average and standard deviation in our proposed scoring metric, with SD. The results are without finetuning. Prompt: ``group of cyclists racing in a scenic countryside''. More examples can be found in the Supplementary Materials.}
  \label{fig:avg_std_qualitative}
\end{figure}

The intention behind this metric is to award higher scores to the latent variables that best preserve output quality. A preliminary visual examination of the image outputs from the pruned models, prior to retraining, provides useful insight into the effectiveness of different approaches. This is evidenced in Fig.~\ref{fig:avg_std_qualitative}, where we illustrate the impact of various methods—summation, multiplication, average only, and standard deviation only—on the visual characteristics of pruned models. Our observations reveal that the `average only' method allows for more pruning before degradation to noise, whereas the `standard deviation only' method tends to preserve sharper features under low compression. When these methods are combined via summation or multiplication, we observe a balance of these attributes, resulting in outcomes that blend both qualities.

\begin{figure*}[thb!]
  \centering
  \begin{subfigure}{0.47\textwidth}
    \centering
    \includegraphics[width=1\linewidth]{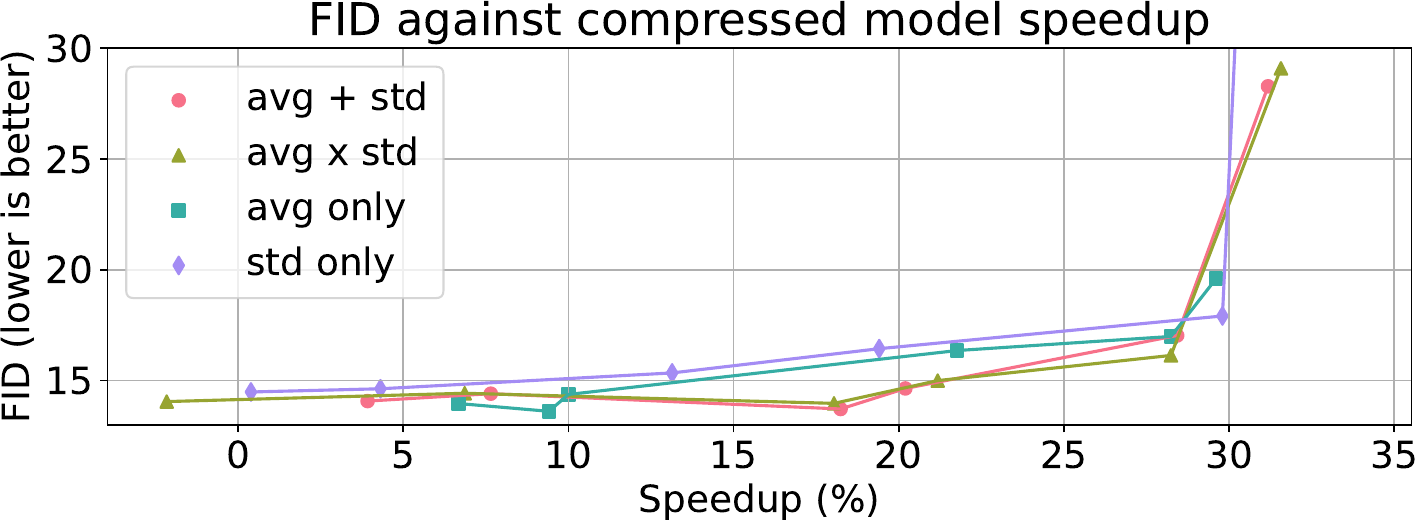}
    \caption{Comparison of the FID of the finetuned compressed model against the speed improvement over the original model}
    \label{fig:avg_std_quantitative_1}
  \end{subfigure}
  \hfill
  \begin{subfigure}{0.47\textwidth}
    \centering
    \includegraphics[width=1\linewidth]{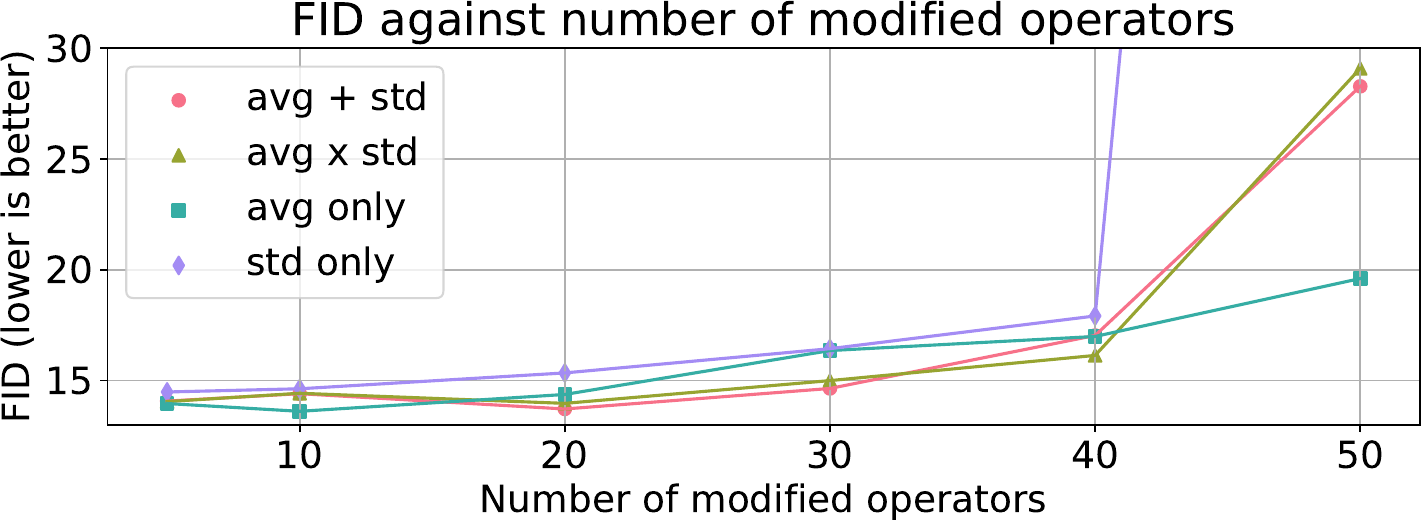}
    \caption{Comparison of the FID of the finetuned compressed model against the number of modified operators}
    \label{fig:avg_std_quantitative_2}
  \end{subfigure}
  \caption{Quantitative comparison of the impact of various combination methods for average and standard deviation in our proposed scoring metric, with UIG. The FID is measured after 20k iterations of finetuning.}
  \label{fig:avg_std_quantitative}
\end{figure*}

A more detailed examination, which compares the FID scores of pruned models post-finetuning for the different formulae, is presented in Fig.~\ref{fig:avg_std_quantitative} for further insight. Notably, a direct comparison with Fig.~\ref{fig:avg_std_qualitative} shows a correlation between the observed degradation to noise in the image before retraining and the decline in FID scores after retraining. This connection tends to support the idea that the visual analysis of models before retraining can serve as an early indicator of post-retraining performance.  Interestingly, the `sum' formula consistently achieves the best or near-best FID scores when modifying up to 40 operators. Yet, as the number of modified operators expands significantly, this positive trend does not persist. We speculate that this outcome is due to the compression process neglecting potential inter-dependencies between operators. Thus, as the number of modified operators escalates, so does the probability of inadvertently eliminating all instances of specific, possibly crucial, information. This particular challenge signals a potential direction for improvement in future research.

% \begin{figure}[thb!]
%   \centering
%   \begin{subfigure}{0.49\linewidth}
%     \centering
%     \includegraphics[width=1\linewidth]{figures/avg_std_quantitative_speedup.pdf}
%     \caption{Comparison of the FID of the finetuned compressed model against the speed improvement over the original model}
%     \label{fig:avg_std_quantitative_1}
%   \end{subfigure}
%   \hfill
%   \begin{subfigure}{0.49\linewidth}
%     \centering
%     \includegraphics[width=1\linewidth]{figures/avg_std_quantitative_modif.pdf}
%     \caption{Comparison of the FID of the finetuned compressed model against the number of modified operators}
%     \label{fig:avg_std_quantitative_2}
%   \end{subfigure}
%   \caption{Quantitative comparison of the impact of various combination methods for average and standard deviation in our proposed scoring metric, with UIG. The FID is measured after 20k iterations of finetuning.}
%   \label{fig:avg_std_quantitative}
% \end{figure}

\subsection{Importance of Preserving the Weights}
\label{subsec:weight_preservation}

Training LDMs typically requires extensive computational resources and large datasets, making the process time-consuming and costly. To address this challenge, our proposed method focuses on weight preservation during pruning. In this subsection, we underscore the significance of this approach and provide empirical evidence of its benefits.

Our experiments, documented in Tab.~\ref{tab:weight_preservation}, compare the performance of models trained from scratch to those with preserved weights, under equivalent compression and training conditions. Performance is quantified using the FID score, which provides a measure of the distance between the model-generated and real data distributions.

The results exhibit a pronounced advantage for models with preserved weights, consistently reporting lower FID scores. This signifies a closer match to the actual data distribution, hence superior image generation quality. Notably, the model with preserved weights comes close to the initial model's performance after merely $20,000$ iterations, whereas the model trained from scratch fails to match the FID score of the model with preserved weights at iteration $0$, even after $50,000$ iterations. This stark performance disparity underscores the pivotal role weight preservation plays in the model training process.

\begin{table}[thb!]
    \scriptsize 
    \centering
    \begin{tabular}{l|c|c|c|c|c}
    \toprule
    Number of train steps & 0 & 4k & 12k & 20k & 50k \\
    \midrule
    From Scratch & 389.96 & 344.27 & 312.86 & 293.07 & 210.43 \\
    With Preserved Weight & 145.19 & 18.29 & 16.65 & 15.74 & \textbf{15.43} \\
    \bottomrule
    \end{tabular}
    \caption{FID scores for our compressed model ($31$ operators modified) trained from scratch and with preserved pre-training weights, for UIG on CelebA-HQ $256\times256$. In both case, the exact same training is applied. The FID for the original model is $13.85$. }
    \label{tab:weight_preservation}
\end{table}

\subsection{Speed-Performance Trade-off}
\label{subsec:tradeoff}

The act of model compression inherently introduces a trade-off between computational speed and performance. As we increase the compression rate, the model's performance, measured by FID, gradually deteriorates. This degradation, however, is not linear but exhibits a threshold-like behavior, beyond which the performance sharply deteriorates.

This phenomenon is illustrated in Figure~\ref{fig:avg_std_quantitative_1}, which demonstrates the relationship between the FID score and the percentage of speedup achieved through compression, relative to the initial model in the context of UIG. As the figure shows, the FID score remains relatively steady, hovering around $15$, for compression rates up to around $30$\%. Beyond this point, the FID score abruptly escalates, indicating a significant drop in the quality of generated images.
We observed a similar trend for other tasks, with a thresholds at $42$ modified operators for T2I and $90$ for UAG.

\subsection{Limitations}
\label{subsec:limitations}

Despite the strengths of our method, there are some limitations to acknowledge. Firstly, our approach does not extend to pruning the decoder part of the model, as it operates after the latent space. Consequently, the method is best suited for LDMs, where most computational expenses occur in the U-Net due to the recursive nature of the process. Secondly, the current approach does not account for dependencies between operators, potentially leading to pruning decisions that are not optimal. These limitations present valuable areas for further improvements to our method.
\section{Conclusion}
\label{sec:conclusion}

In this paper, we have presented a novel architecture pruning algorithm for LDMs that is task-agnostic and leverages the latent space to guide the pruning process. An integral part of our approach is the introduction of a new scoring metric that enables the direct comparison of latent representations, providing a robust, quantifiable measure for pruning decisions. Our approach addresses unique challenges posed by generative models, transcending task-specific limitations of existing strategies, and leads to compact models with faster inference speed and fewer parameters, without substantially sacrificing performance.

{
    \small
    \bibliographystyle{ieeenat_fullname}
    \bibliography{main}

\begin{thebibliography}{37}
\providecommand{\natexlab}[1]{#1}
\providecommand{\url}[1]{\texttt{#1}}
\expandafter\ifx\csname urlstyle\endcsname\relax
  \providecommand{\doi}[1]{doi: #1}\else
  \providecommand{\doi}{doi: \begingroup \urlstyle{rm}\Url}\fi

\bibitem[Bolya and Hoffman(2023)]{ToMeSD}
Daniel Bolya and Judy Hoffman.
\newblock Token merging for fast stable diffusion.
\newblock \emph{arXiv}, 2023.

\bibitem[Castells and Yeom(2023)]{AutoBot}
Thibault Castells and Seul-Ki Yeom.
\newblock Automatic neural network pruning that efficiently preserves the model
  accuracy.
\newblock In \emph{2nd International Workshop on Practical Deep Learning in the
  Wild}, 2023.

\bibitem[CompVis(2023)]{SD_training}
CompVis.
\newblock Stable diffusion training.
\newblock \url{https://huggingface.co/CompVis/stable-diffusion-v1-4}, 2023.

\bibitem[Dargavel~Smith(2022)]{audiodiffusion}
Robert Dargavel~Smith.
\newblock Audiodiffusion.
\newblock \url{https://huggingface.co/teticio/latent-audio-diffusion-256},
  2022.

\bibitem[Ding et~al.(2021)Ding, Yang, Hong, Zheng, Zhou, Yin, Lin, Zou, Shao,
  Yang, and Tang]{cogview}
Ming Ding, Zhuoyi Yang, Wenyi Hong, Wendi Zheng, Chang Zhou, Da Yin, Junyang
  Lin, Xu Zou, Zhou Shao, Hongxia Yang, and Jie Tang.
\newblock Cogview: Mastering text-to-image generation via transformers.
\newblock In \emph{Advances in Neural Information Processing Systems
  (NeurIPS)}, 2021.

\bibitem[Ding et~al.(2022)Ding, Zheng, Hong, and Tang]{cogview2}
Ming Ding, Wendi Zheng, Wenyi Hong, and Jie Tang.
\newblock Cogview2: Faster and better text-to-image generation via hierarchical
  transformers.
\newblock In \emph{Advances in Neural Information Processing Systems
  (NeurIPS)}, 2022.

\bibitem[Fang et~al.(2023)Fang, Ma, Song, Mi, and Wang]{fang2023depgraph}
Gongfan Fang, Xinyin Ma, Mingli Song, Michael~Bi Mi, and Xinchao Wang.
\newblock Depgraph: Towards any structural pruning.
\newblock In \emph{{IEEE/CVF} Conference on Computer Vision and Pattern
  Recognition {(CVPR)}}, 2023.

\bibitem[Fernandez et~al.(2023)Fernandez, Kahn, Na, Bisk, and
  Strubell]{fernandez2023framework}
Jared Fernandez, Jacob Kahn, Clara Na, Yonatan Bisk, and Emma Strubell.
\newblock The framework tax: Disparities between inference efficiency in
  research and deployment, 2023.

\bibitem[Gafni et~al.(2022)Gafni, Polyak, Ashual, Sheynin, Parikh, and
  Taigman]{makeascene}
Oran Gafni, Adam Polyak, Oron Ashual, Shelly Sheynin, Devi Parikh, and Yaniv
  Taigman.
\newblock Make-a-scene: Scene-based text-to-image generation with human priors.
\newblock In \emph{European Conference on Computer Vision {(ECCV)}}, 2022.

\bibitem[Goodfellow et~al.(2014)Goodfellow, Pouget-Abadie, Mirza, Xu,
  Warde-Farley, Ozair, Courville, and Bengio]{gans}
Ian Goodfellow, Jean Pouget-Abadie, Mehdi Mirza, Bing Xu, David Warde-Farley,
  Sherjil Ozair, Aaron Courville, and Yoshua Bengio.
\newblock Generative adversarial nets.
\newblock In \emph{Advances in Neural Information Processing Systems
  (NeurIPS)}, 2014.

\bibitem[Han et~al.(2023)Han, Li, Zhang, Milanfar, Metaxas, and Yang]{svdiff}
Ligong Han, Yinxiao Li, Han Zhang, Peyman Milanfar, Dimitris Metaxas, and Feng
  Yang.
\newblock Svdiff: Compact parameter space for diffusion fine-tuning.
\newblock \emph{arXiv preprint arXiv:2303.11305}, 2023.

\bibitem[Heusel et~al.(2017)Heusel, Ramsauer, Unterthiner, Nessler, and
  Hochreiter]{FID}
Martin Heusel, Hubert Ramsauer, Thomas Unterthiner, Bernhard Nessler, and Sepp
  Hochreiter.
\newblock Gans trained by a two time-scale update rule converge to a local nash
  equilibrium.
\newblock In \emph{Advances in Neural Information Processing Systems
  (NeurIPS)}, 2017.

\bibitem[Ho et~al.(2020)Ho, Jain, and Abbeel]{DDPM}
Jonathan Ho, Ajay Jain, and Pieter Abbeel.
\newblock Denoising diffusion probabilistic models.
\newblock In \emph{Advances in Neural Information Processing Systems
  (NeurIPS)}, 2020.

\bibitem[Karras et~al.(2018)Karras, Aila, Laine, and Lehtinen]{celeba_hq}
Tero Karras, Timo Aila, Samuli Laine, and Jaakko Lehtinen.
\newblock Progressive growing of {GAN}s for improved quality, stability, and
  variation, 2018.

\bibitem[Kilgour et~al.(2019)Kilgour, Zuluaga, Roblek, and Sharifi]{FAD}
Kevin Kilgour, Mauricio Zuluaga, Dominik Roblek, and Matthew Sharifi.
\newblock Fr\'echet audio distance: A metric for evaluating music enhancement
  algorithms.
\newblock \emph{arXiv preprint arXiv:1812.08466}, 2019.

\bibitem[Kim et~al.(2023)Kim, Song, Castells, and Choi]{BKsdm}
Bo-Kyeong Kim, Hyoung-Kyu Song, Thibault Castells, and Shinkook Choi.
\newblock Bk-sdm: A lightweight, fast, and cheap version of stable diffusion.
\newblock \emph{arXiv}, 2023.

\bibitem[Kingma and Welling(2014)]{vae}
Diederik~P. Kingma and Max Welling.
\newblock {Auto-Encoding Variational Bayes}.
\newblock In \emph{International Conference on Learning Representations
  {(ICLR)}}, 2014.

\bibitem[Li et~al.(2021)Li, Yu, Fu, Zhang, Zhao, You, Yu, Wang, and
  Lin~(Celine)]{li2021hwnasbench}
Chaojian Li, Zhongzhi Yu, Yonggan Fu, Yongan Zhang, Yang Zhao, Haoran You,
  Qixuan Yu, Yue Wang, and Yingyan Lin~(Celine).
\newblock Hw-nas-bench: Hardware-aware neural architecture search benchmark.
\newblock In \emph{International Conference on Learning Representations
  {(ICLR)}}, 2021.

\bibitem[Li et~al.(2017)Li, Kadav, Durdanovic, Samet, and Graf]{l1norm}
Hao Li, Asim Kadav, Igor Durdanovic, Hanan Samet, and Hans~Peter Graf.
\newblock Pruning filters for efficient convnets.
\newblock In \emph{International Conference on Learning Representations
  {(ICLR)}}, 2017.

\bibitem[Li et~al.(2023{\natexlab{a}})Li, Lian, Liu, Yang, Dong, Kang, Zhang,
  and Keutzer]{li2023qdiff}
Xiuyu Li, Long Lian, Yijiang Liu, Huanrui Yang, Zhen Dong, Daniel Kang,
  Shanghang Zhang, and Kurt Keutzer.
\newblock Q-diffusion: Quantizing diffusion models.
\newblock \emph{arXiv preprint arXiv:2302.04304}, 2023{\natexlab{a}}.

\bibitem[Li et~al.(2023{\natexlab{b}})Li, Wang, Jin, Hu, Chemerys, Fu, Wang,
  Tulyakov, and Ren]{snapfusion}
Yanyu Li, Huan Wang, Qing Jin, Ju Hu, Pavlo Chemerys, Yun Fu, Yanzhi Wang,
  Sergey Tulyakov, and Jian Ren.
\newblock Snapfusion: Text-to-image diffusion model on mobile devices within
  two seconds.
\newblock In \emph{Advances in Neural Information Processing Systems
  (NeurIPS)}, 2023{\natexlab{b}}.

\bibitem[Lin et~al.(2014)Lin, Maire, Belongie, Hays, Perona, Ramanan,
  Doll{\'{a}}r, and Zitnick]{COCO}
Tsung{-}Yi Lin, Michael Maire, Serge~J. Belongie, James Hays, Pietro Perona,
  Deva Ramanan, Piotr Doll{\'{a}}r, and C.~Lawrence Zitnick.
\newblock Microsoft {COCO:} common objects in context.
\newblock In \emph{European Conference on Computer Vision {(ECCV)}}, 2014.

\bibitem[Liu et~al.(2023)Liu, Chen, Yuan, Mei, Liu, Mandic, Wang, and
  Plumbley]{liu2023audioldm}
Haohe Liu, Zehua Chen, Yi Yuan, Xinhao Mei, Xubo Liu, Danilo Mandic, Wenwu
  Wang, and Mark~D Plumbley.
\newblock Audioldm: Text-to-audio generation with latent diffusion models.
\newblock \emph{arXiv preprint arXiv:2301.12503}, 2023.

\bibitem[Liu et~al.(2017)Liu, Li, Shen, Huang, Yan, and
  Zhang]{liunetworkslimming}
Zhuang Liu, Jianguo Li, Zhiqiang Shen, Gao Huang, Shoumeng Yan, and Changshui
  Zhang.
\newblock Learning efficient convolutional networks through network slimming.
\newblock In \emph{{IEEE} International Conference on Computer Vision
  {(ICCV)}}, 2017.

\bibitem[OFA-Sys(2022)]{OFA_sdm}
OFA-Sys.
\newblock Small stable diffusion.
\newblock \url{https://huggingface.co/OFA-Sys/small-stable-diffusion-v0}, 2022.

\bibitem[Pernias et~al.(2024)Pernias, Rampas, Leon~Richter, Pal, and
  Aubreville]{wurstchen_v2}
Pablo Pernias, Dominic Rampas, Mats Leon~Richter, Christopher Pal, and Marc
  Aubreville.
\newblock W\"urstchen: An efficient architecture for large-scale text-to-image
  diffusion models.
\newblock In \emph{International Conference on Learning Representations
  {(ICLR)}}, 2024.

\bibitem[Qualcomm(2020)]{aimet}
Qualcomm.
\newblock Aimet.
\newblock \url{https://github.com/quic/aimet}, 2020.

\bibitem[Quinn~Nichol et~al.(2022)Quinn~Nichol, Dhariwal, Ramesh, Shyam,
  Mishkin, McGrew, Sutskever, and Chen]{glide}
Alexander Quinn~Nichol, Prafulla Dhariwal, Aditya Ramesh, Pranav Shyam, Pamela
  Mishkin, Bob McGrew, Ilya Sutskever, and Mark Chen.
\newblock {GLIDE:} towards photorealistic image generation and editing with
  text-guided diffusion models.
\newblock In \emph{International Conference on Machine Learning {(ICML)}},
  2022.

\bibitem[Ramesh et~al.(2021)Ramesh, Pavlov, Goh, Gray, Voss, Radford, Chen, and
  Sutskever]{dalle}
Aditya Ramesh, Mikhail Pavlov, Gabriel Goh, Scott Gray, Chelsea Voss, Alec
  Radford, Mark Chen, and Ilya Sutskever.
\newblock Zero-shot text-to-image generation.
\newblock In \emph{International Conference on Machine Learning {(ICML)}},
  2021.

\bibitem[Ramesh et~al.(2022)Ramesh, Dhariwal, Nichol, Chu, and Chen]{dalle2}
Aditya Ramesh, Prafulla Dhariwal, Alex Nichol, Casey Chu, and Mark Chen.
\newblock Hierarchical text-conditional image generation with clip latents,
  2022.

\bibitem[Rombach et~al.(2021)Rombach, Blattmann, Lorenz, Esser, and
  Ommer]{rombach2021highresolution}
Robin Rombach, Andreas Blattmann, Dominik Lorenz, Patrick Esser, and Björn
  Ommer.
\newblock High-resolution image synthesis with latent diffusion models.
\newblock In \emph{{IEEE/CVF} Conference on Computer Vision and Pattern
  Recognition {(CVPR)}}, 2021.

\bibitem[Saharia et~al.(2022)Saharia, Chan, Saxena, Li, Whang, Denton,
  Ghasemipour, Gontijo~Lopes, Karagol~Ayan, Salimans, Ho, Fleet, and
  Norouzi]{imagen}
Chitwan Saharia, William Chan, Saurabh Saxena, Lala Li, Jay Whang, Emily~L
  Denton, Kamyar Ghasemipour, Raphael Gontijo~Lopes, Burcu Karagol~Ayan, Tim
  Salimans, Jonathan Ho, David~J Fleet, and Mohammad Norouzi.
\newblock Photorealistic text-to-image diffusion models with deep language
  understanding.
\newblock In \emph{Advances in Neural Information Processing Systems
  (NeurIPS)}, 2022.

\bibitem[Salimans et~al.(2016)Salimans, Goodfellow, Zaremba, Cheung, Radford,
  Chen, and Chen]{IS}
Tim Salimans, Ian Goodfellow, Wojciech Zaremba, Vicki Cheung, Alec Radford, Xi
  Chen, and Xi Chen.
\newblock Improved techniques for training gans.
\newblock In \emph{Advances in Neural Information Processing Systems
  (NeurIPS)}, 2016.

\bibitem[Schuhmann and Beaumont(2023)]{laion_aesthetics}
Christoph Schuhmann and Romain Beaumont.
\newblock Laion-aesthetics, 2023.

\bibitem[Tao et~al.(2023)Tao, Bao, Tang, and Xu]{galip}
Ming Tao, Bing-Kun Bao, Hao Tang, and Changsheng Xu.
\newblock Galip: Generative adversarial clips for text-to-image synthesis.
\newblock In \emph{{IEEE/CVF} Conference on Computer Vision and Pattern
  Recognition {(CVPR)}}, 2023.

\bibitem[Zeng et~al.(2022)Zeng, Vahdat, Williams, Gojcic, Litany, Fidler, and
  Kreis]{zeng2022lion}
Xiaohui Zeng, Arash Vahdat, Francis Williams, Zan Gojcic, Or Litany, Sanja
  Fidler, and Karsten Kreis.
\newblock Lion: Latent point diffusion models for 3d shape generation.
\newblock In \emph{Advances in Neural Information Processing Systems
  (NeurIPS)}, 2022.

\bibitem[Zhou et~al.(2022)Zhou, Zhang, Chen, Li, Tensmeyer, Yu, Gu, Xu, and
  Sun]{lafite}
Yufan Zhou, Ruiyi Zhang, Changyou Chen, Chunyuan Li, Chris Tensmeyer, Tong Yu,
  Jiuxiang Gu, Jinhui Xu, and Tong Sun.
\newblock Lafite: Towards language-free training for text-to-image generation.
\newblock In \emph{{IEEE/CVF} Conference on Computer Vision and Pattern
  Recognition {(CVPR)}}, 2022.

\end{thebibliography}
}

\clearpage
\setcounter{page}{1}
\maketitlesupplementary

\section{Operator Modification}
\label{sec:oprem}

\begin{figure*}[ht]
  \centering
  \includegraphics[width=0.8\linewidth]{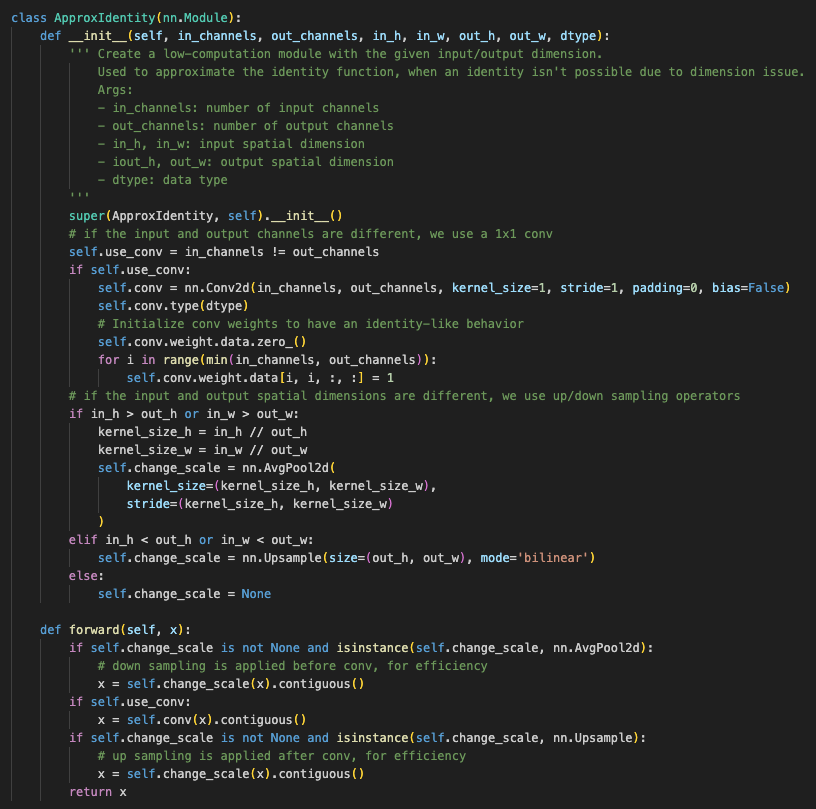}
  \caption{Code used to replace the operators with different input/output dimension (typically, a convolution).}
  \label{fig:code}
\end{figure*}

Fig.~\ref{fig:code} shows the code used to replace the operators with different input and output number of channels and/or spatial dimensions.

%----------------------------------------------------------------------------------------------------------
%----------------------------------------------------------------------------------------------------------

\section{Hyper-parameters}
\label{sec:hyperparams}

In Tab.~\ref{tab:hparams}, we give the hyper-parameters used for our experiments on Text-to-Image generation (T2I), Unconditional Image Generation (UIG) and Unconditional Audio Generation (UAG). `feat. KD coef.' and `out KD coef.' refer to the coefficient used in the knowledge-distillation loss applied at the feature-level and output-level, respectively.

\begin{table*}[ht]
    \centering
    \small
    \begin{tabular}{lcccccc}
        \toprule
        Task & lr & batch size & gradient accumulation & iterations & feat. KD coef. & out KD coef. \\
        \midrule
        T2I Generation & $3e^{-5}$ & 64 & 4 & 50,000 & 0.7 & 0.7 \\
        UIG & $5e^{-6}$ & 32 & 4 & 50,000 & 300 & 300 \\
        UAG & $1e^{-4}$ & 64 & 2 & 12,000 & 10 & 10 \\
        \bottomrule
    \end{tabular}
    \caption{Hyper-parameters used in our experiments. We used the same hyper-parameters for all compression.}
    \label{tab:hparams}
\end{table*}

%----------------------------------------------------------------------------------------------------------
%----------------------------------------------------------------------------------------------------------

\section{Scoring Metric Composition: more results}

\begin{figure*}[thb!]
  \centering
  \includegraphics[width=1\linewidth]{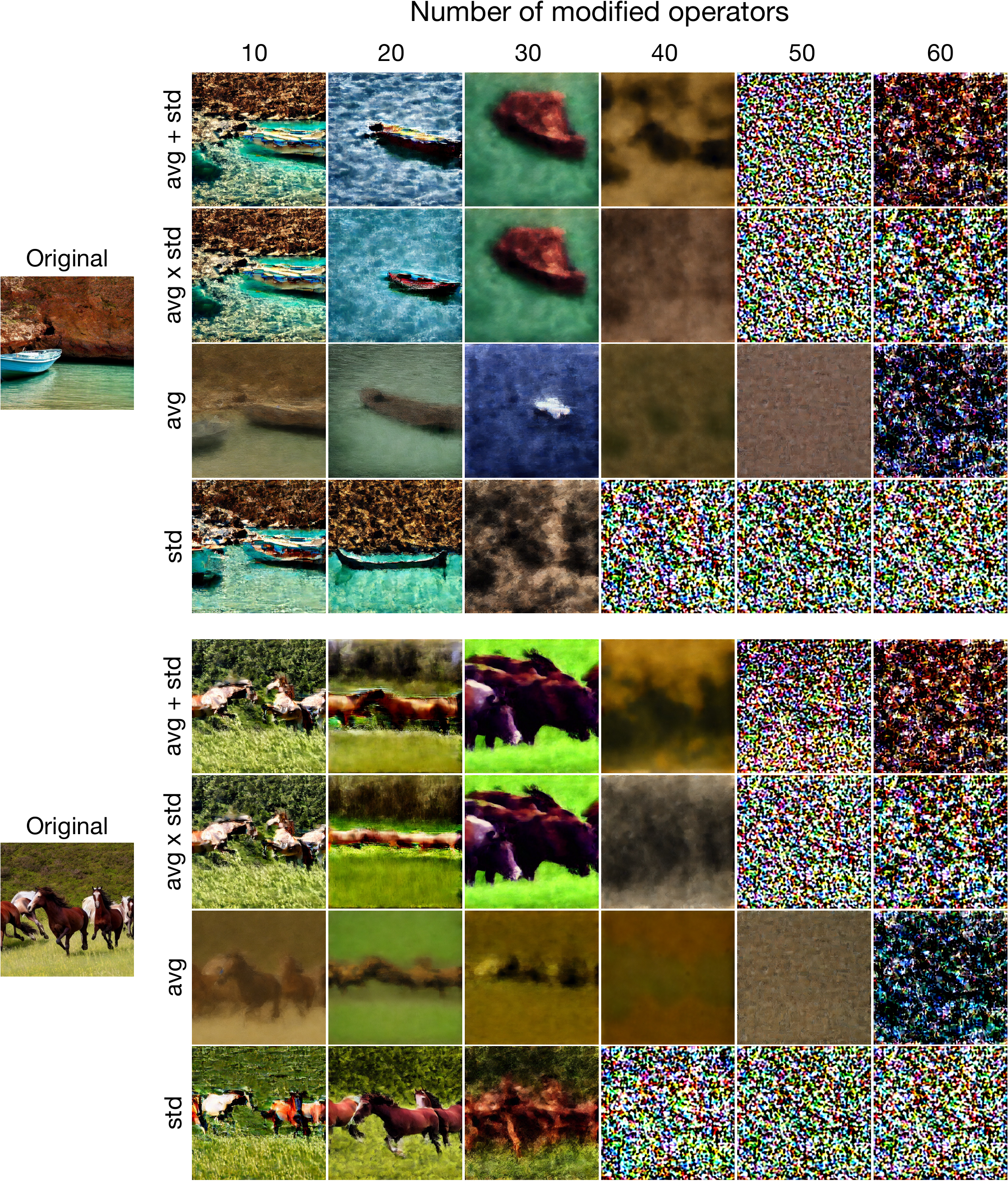}
  \caption{Qualitative comparison of the impact of various combination methods for average and standard deviation in our proposed scoring metric, with SD. The results are without finetuning. Prompts: ``boat on a beautiful sea'', ``group of wild horses galloping through a meadow''.}
  \label{fig:avg_std_qualitative_supp}
\end{figure*}

In Fig.~\ref{fig:avg_std_qualitative_supp}, we provide additional qualitative comparison examples for the scoring metric composition comparison on T2I with SD. The results are without finetuning.

%----------------------------------------------------------------------------------------------------------
%----------------------------------------------------------------------------------------------------------

\section{Importance Score vs Block}
\label{sec:importance_score}

\begin{figure*}[thb!]
  \centering
  \includegraphics[width=1.0\linewidth]{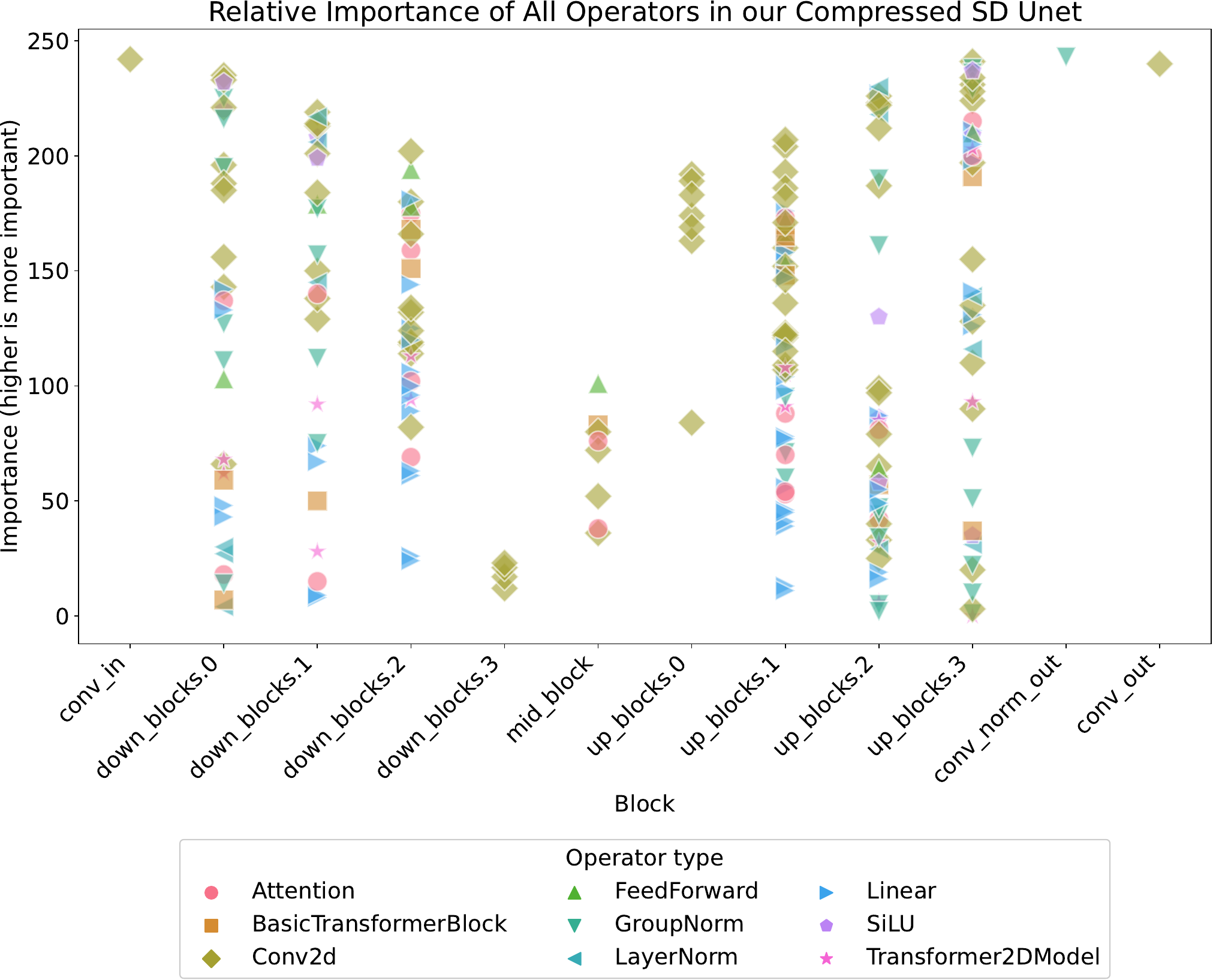}
  \caption{Importance ranking of the operators in SD Unet, as determined by LDPruner. Lower values indicate less importance to the Unet output.}
  \label{fig:importance_score_T2I}
\end{figure*}

\begin{figure*}[thb!]
  \centering
  \includegraphics[width=1.0\linewidth]{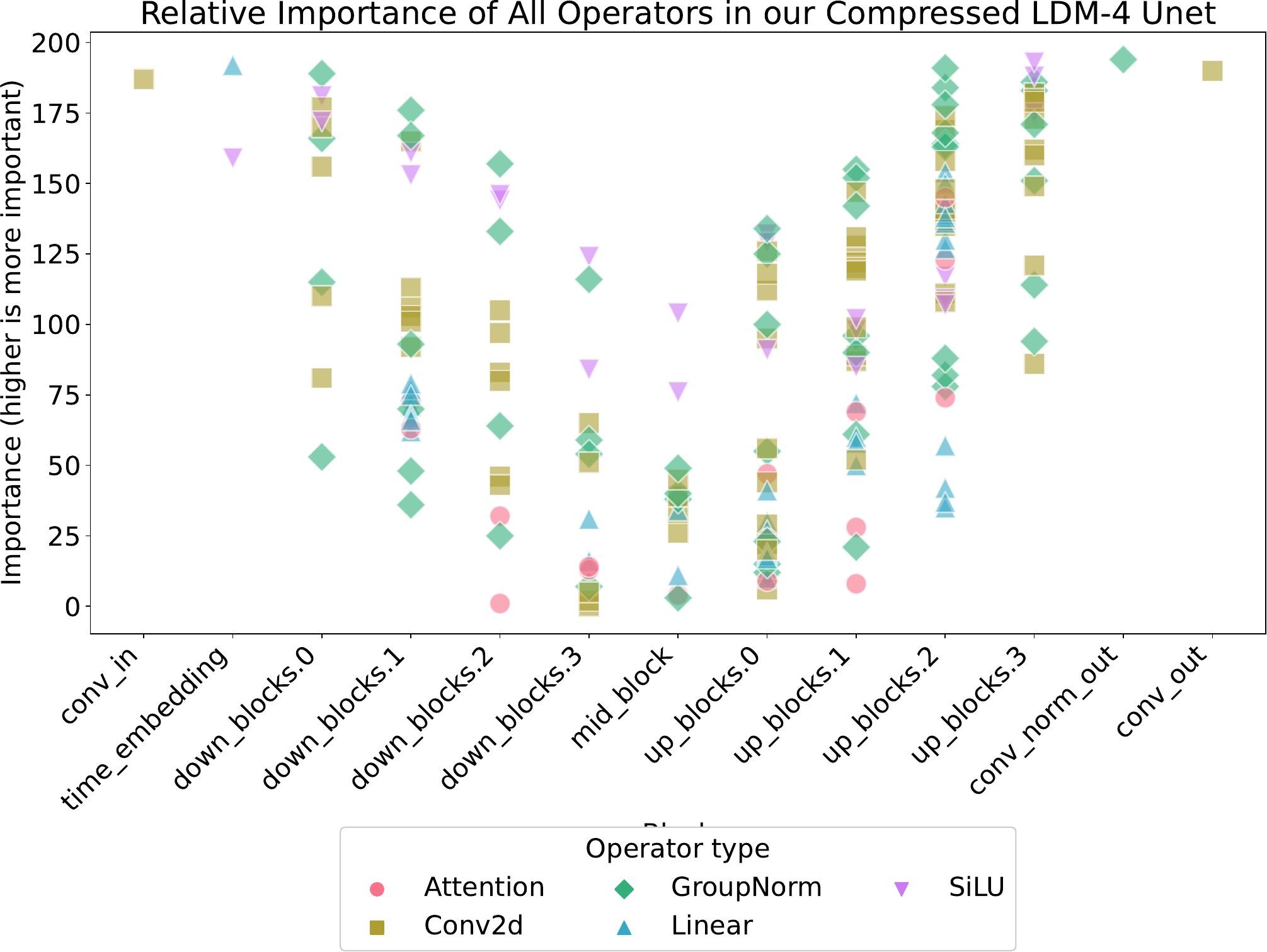}
  \caption{Importance ranking of the operators in LDM-4 Unet, as determined by LDPruner. Lower values indicate less importance to the Unet output.}
  \label{fig:importance_score_UIG}
\end{figure*}

\begin{figure*}[thb!]
  \centering
  \includegraphics[width=1.0\linewidth]{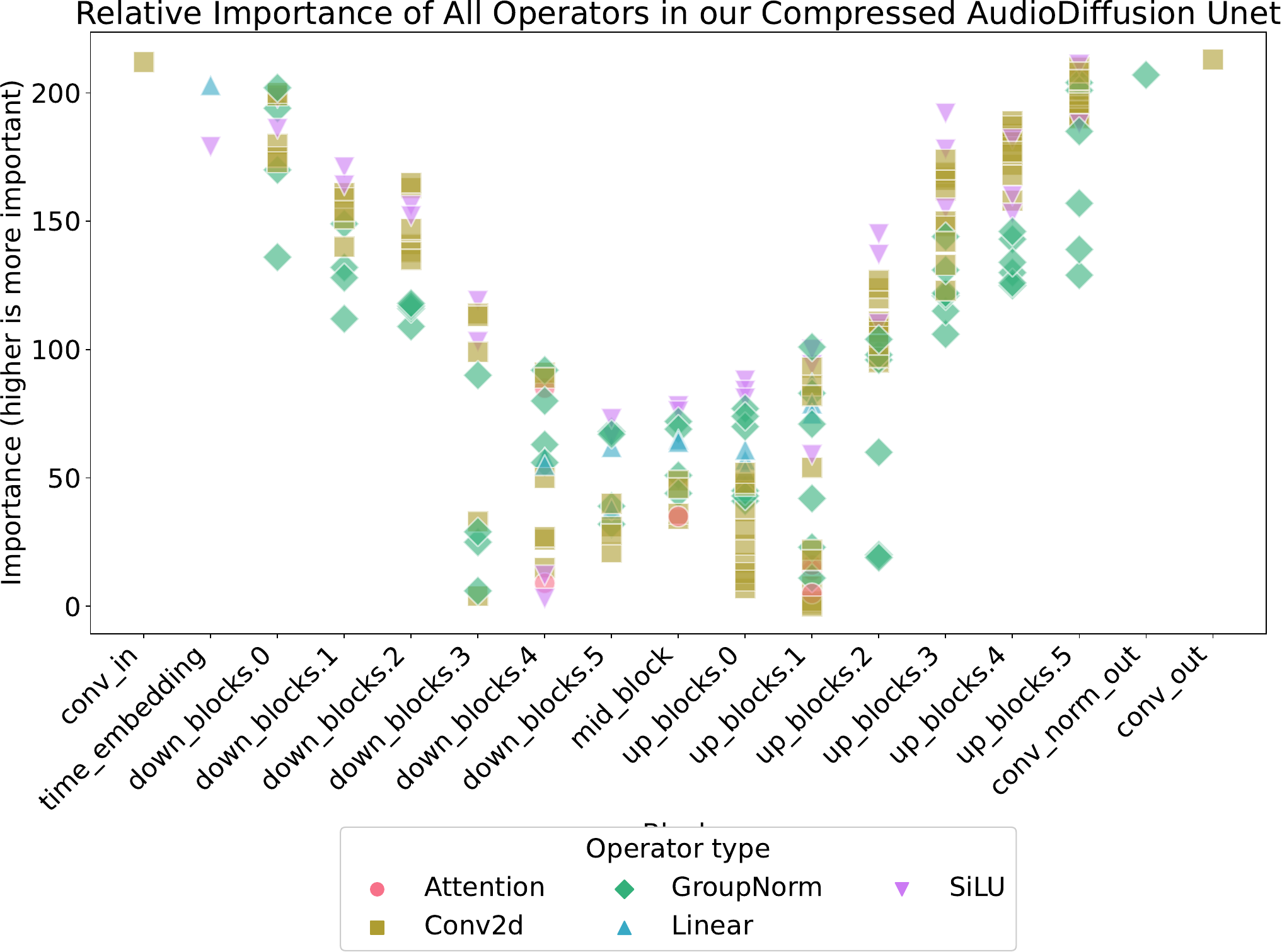}
  \caption{Importance ranking of the operators in AudioDiffusion Unet, as determined by LDPruner. Lower values indicate less importance to the Unet output.}
  \label{fig:importance_score_UAG}
\end{figure*}

In Figs.~\ref{fig:importance_score_T2I},~\ref{fig:importance_score_UIG} and~\ref{fig:importance_score_UAG} , we show the relative importance of each operator in the Unet of SD, LDM-4 and AudioDiffusion, respectively. These scores provide insights into the relative contribution of individual operators to the overall models. The relationship between the BasicTransformerBlock, Transformer2D and Attention operators is illustrated in Fig.~\ref{fig:transformer2d}.

%----------------------------------------------------------------------------------------------------------
%----------------------------------------------------------------------------------------------------------

\section{Visualization of Modified Operator Distribution}
\label{sec:modif_op_distrib}

\begin{figure*}[thb!]
  \centering
  \includegraphics[width=1.0\linewidth]{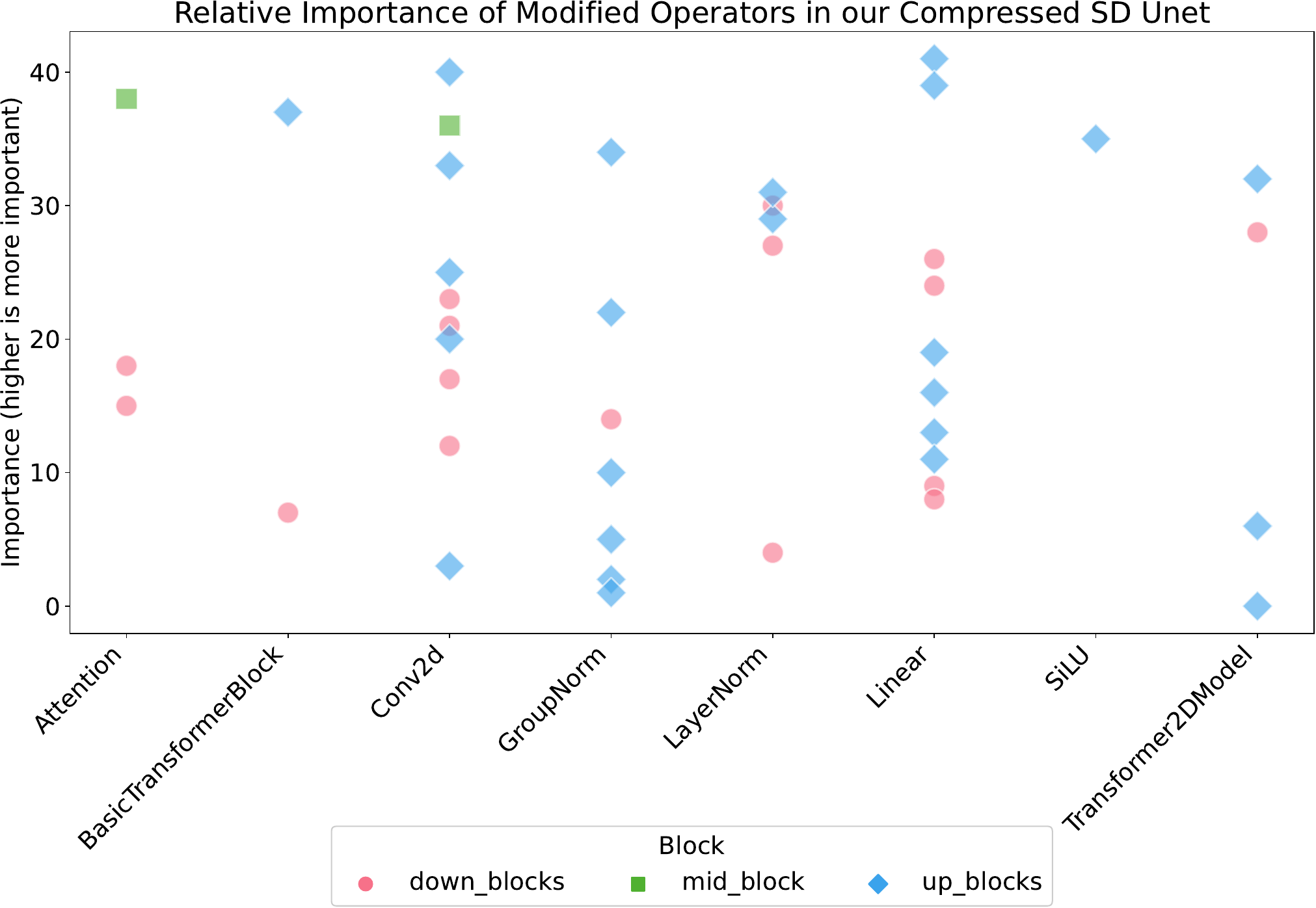}
  \caption{Importance ranking of the modified operators in SD Unet, as determined by LDPruner. Lower values indicate less importance to the Unet output.}
  \label{fig:T2I_operator_type_vs_rank}
\end{figure*}

\begin{figure*}[thb!]
  \centering
  \includegraphics[width=1.0\linewidth]{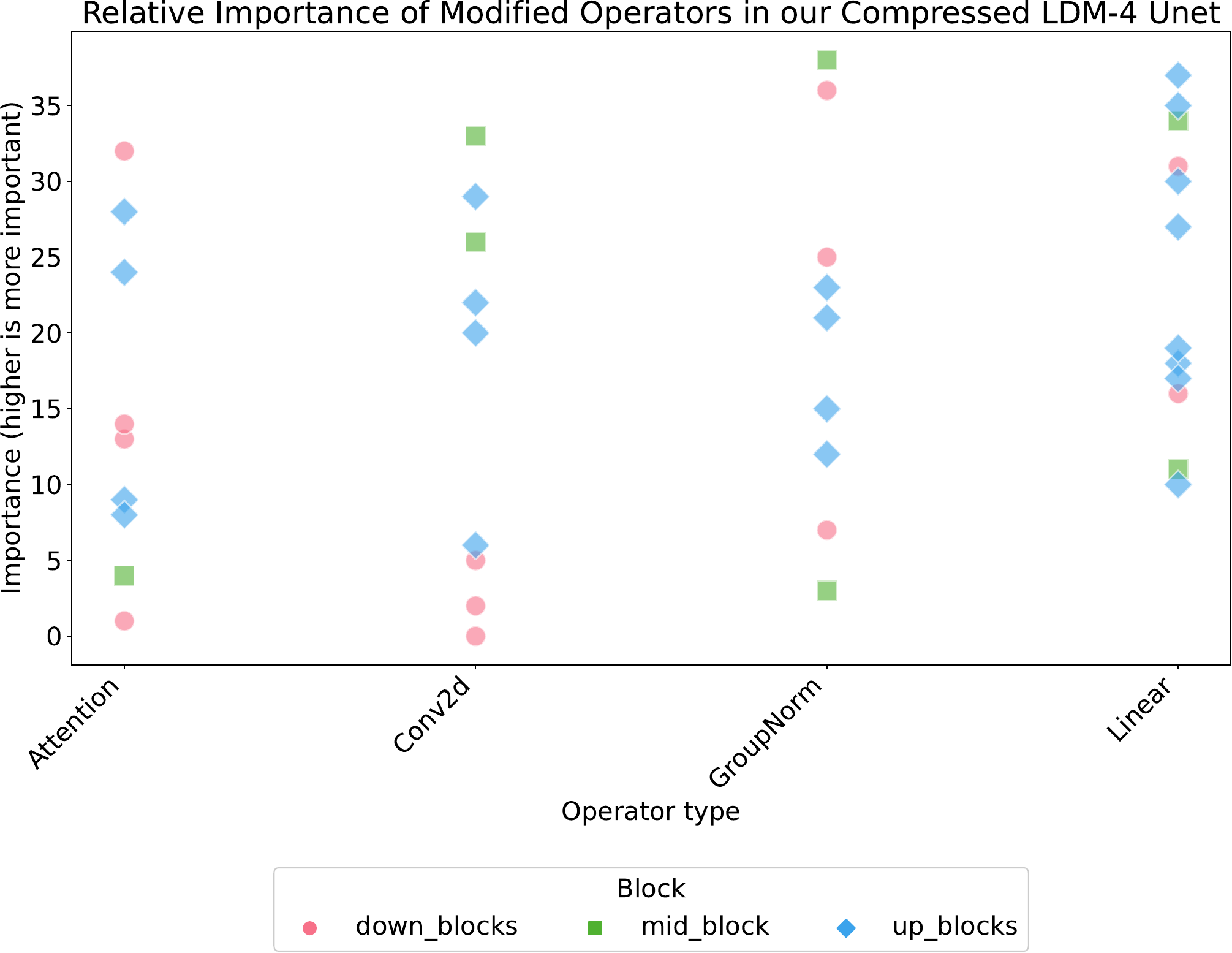}
  \caption{Importance ranking of the modified operators in LDM-4 Unet, as determined by LDPruner. Lower values indicate less importance to the Unet output.}
  \label{fig:UIG_operator_type_vs_rank}
\end{figure*}

\begin{figure*}[thb!]
  \centering
  \includegraphics[width=1.0\linewidth]{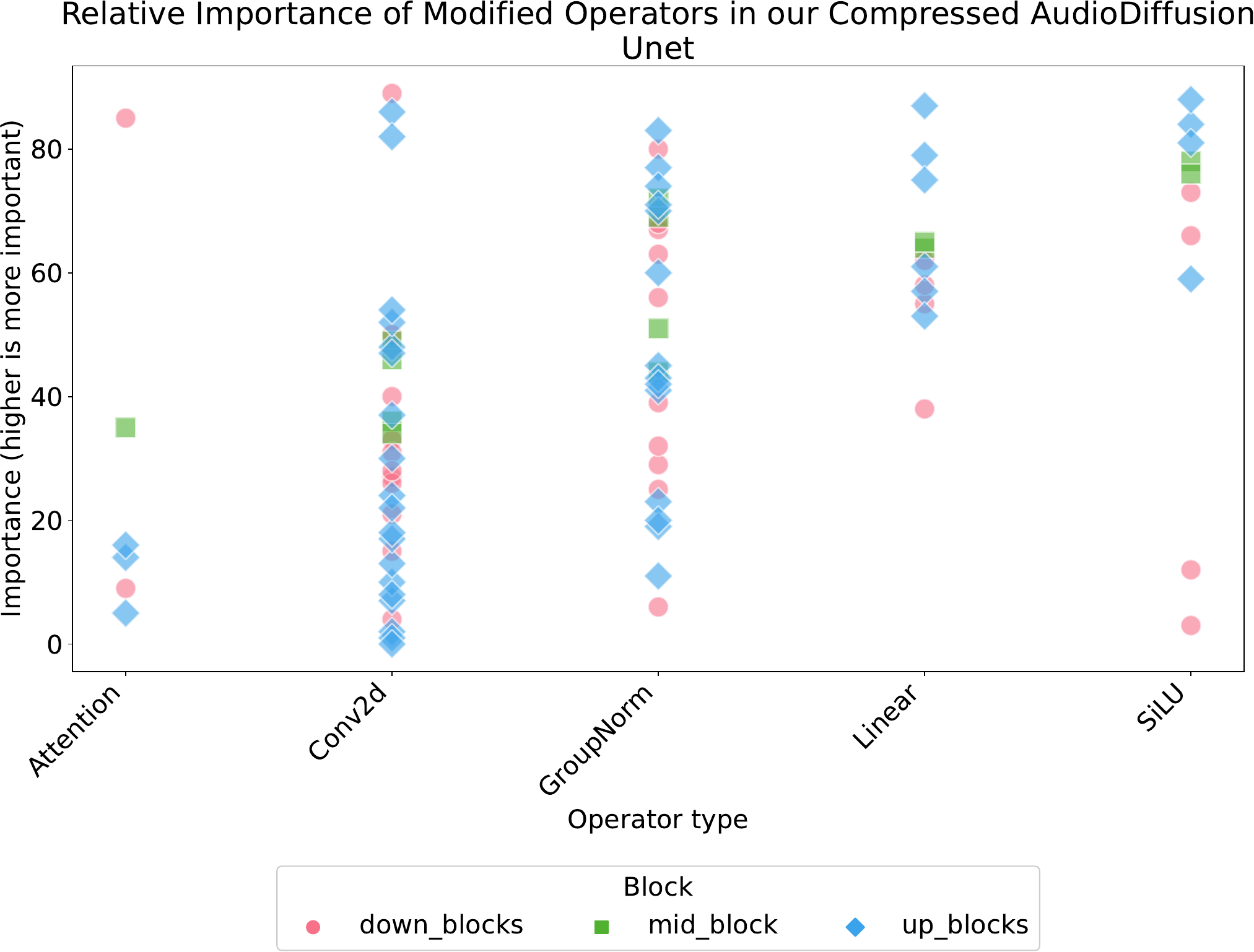}
  \caption{Importance ranking of the modified operators in AudioDiffusion Unet, as determined by LDPruner. Lower values indicate less importance to the Unet output.}
  \label{fig:UAG_operator_type_vs_rank}
\end{figure*}

\begin{figure*}[thb!]
  \centering
  \includegraphics[width=1.0\linewidth]{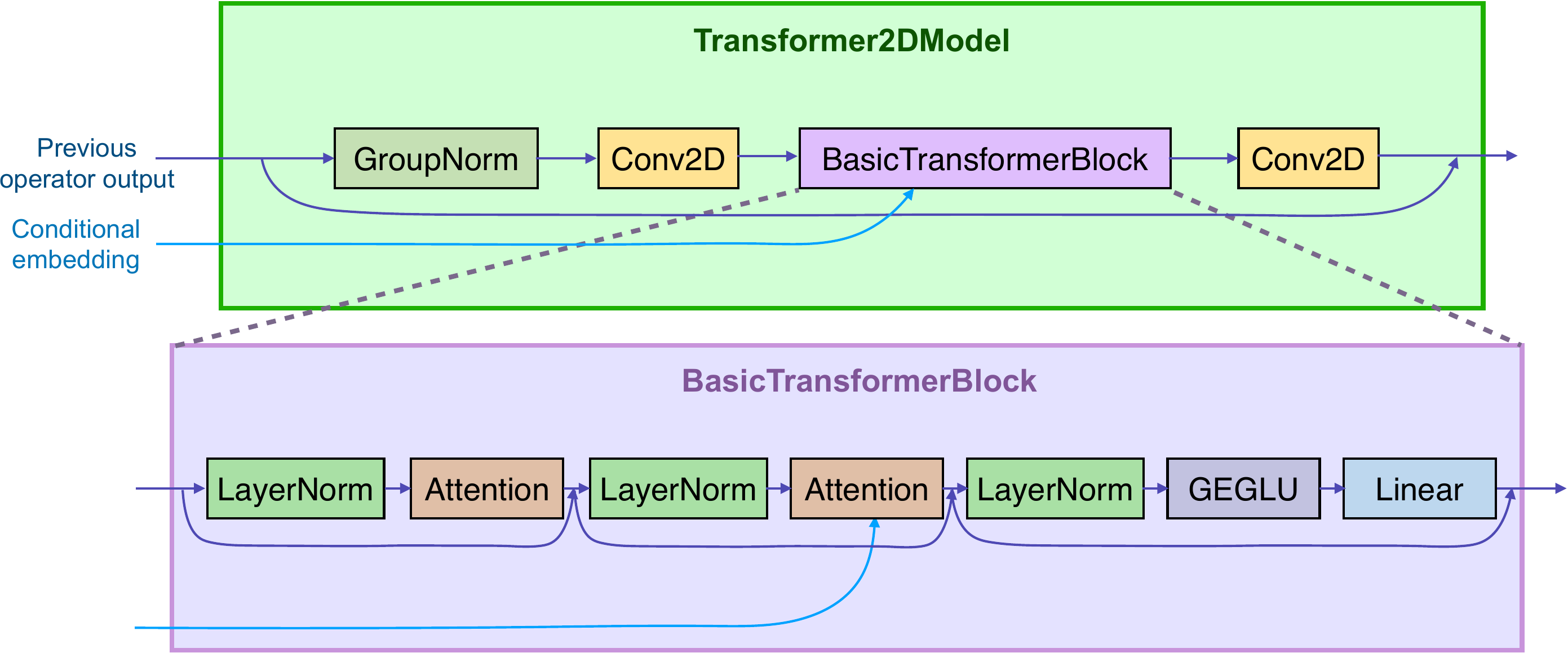}
  \caption{A simplified view of the Transformer2DModel operator, illustrating the relationship between the Transformer2DModel, BasicTransformerBlock, and Attention operators.}
  \label{fig:transformer2d}
\end{figure*}

In Figs.~\ref{fig:T2I_operator_type_vs_rank},~\ref{fig:UIG_operator_type_vs_rank}, and~\ref{fig:UAG_operator_type_vs_rank}, we visualize the distribution by type of the modified operators within the Unet structures of SD, LDM-4, and AudioDiffusion, respectively. The operators are categorized by their type and the block they inhabit. The relationship between the BasicTransformerBlock, Transformer2D and Attention operators is illustrated in Fig.~\ref{fig:transformer2d}.

%----------------------------------------------------------------------------------------------------------
%----------------------------------------------------------------------------------------------------------

\end{document}